\documentclass[runningheads]{llncs}

\usepackage[final,year=2024,ID=31]{eccv}
\usepackage{eccvabbrv}

\usepackage{graphicx}
\usepackage{booktabs}
\usepackage{color}
\usepackage{colortbl}
\usepackage{array}

\usepackage[accsupp]{axessibility}
\usepackage{pifont}
\usepackage{tabularx}
\usepackage{algorithm}
\usepackage{algpseudocode}
\usepackage{multirow}
\usepackage{tikz}
\usepackage{soul}

\definecolor{cvprblue}{rgb}{0.21,0.49,0.74}
\definecolor{lightgreen}{rgb}{0.,0.5,0.}
\definecolor{lightred}{rgb}{0.5,0.,0.}
\definecolor{linkcolor}{rgb}{0.956,0.298,0.235} 
\usepackage[pagebackref,breaklinks,colorlinks,citecolor=cvprblue, linkcolor=linkcolor]{hyperref}

\usepackage{xspace}
\newcommand{\ours}{SOLE\xspace}
\usepackage{orcidlink}
\renewcommand{\arraystretch}{1.25}

\begin{document}

\title{Segment Any 3D Object with Language}

\titlerunning{\ours}

\author{
Seungjun Lee
}
\author{Seungjun Lee\thanks{Equal Contribution.}\inst{1} \and Yuyang Zhao$^\star$\inst{2} \and Gim Hee Lee\inst{2}\index{Lee, Gim Hee}}
\authorrunning{S. Lee et al.}
\institute{Korea University \and National University of Singapore \\
\normalsize\url{https://cvrp-sole.github.io}}

\maketitle
\vspace{-15pt}
\begin{figure}[h]
\captionsetup[subfigure]{}
  \centering
  \begin{subfigure}{0.3233\linewidth}\includegraphics[width=1.0\linewidth]{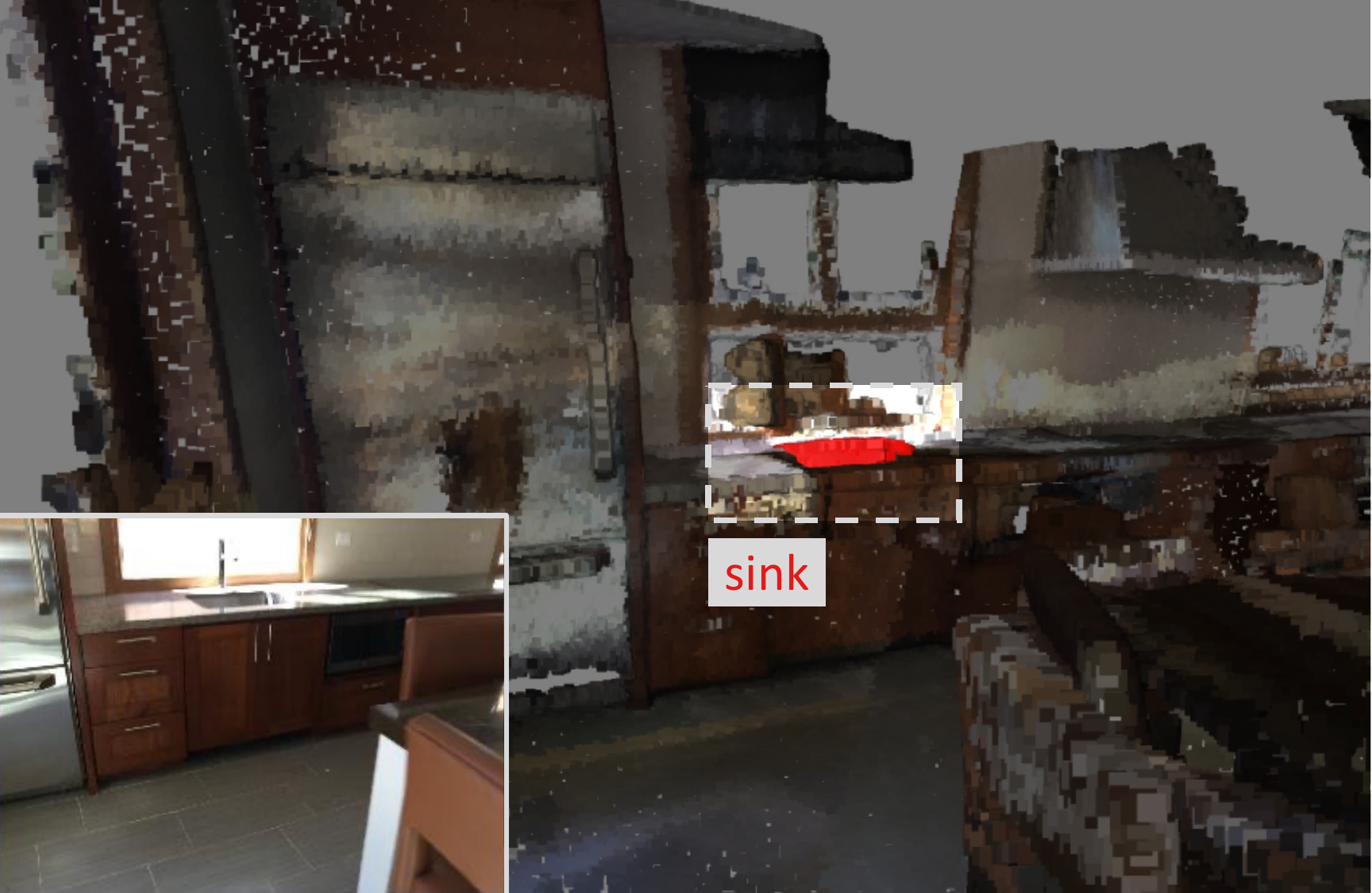}
    \caption{\textit{``Can I wash my hands?''}}
  \end{subfigure}
  \begin{subfigure}{0.3233\linewidth}\includegraphics[width=1.0\linewidth]{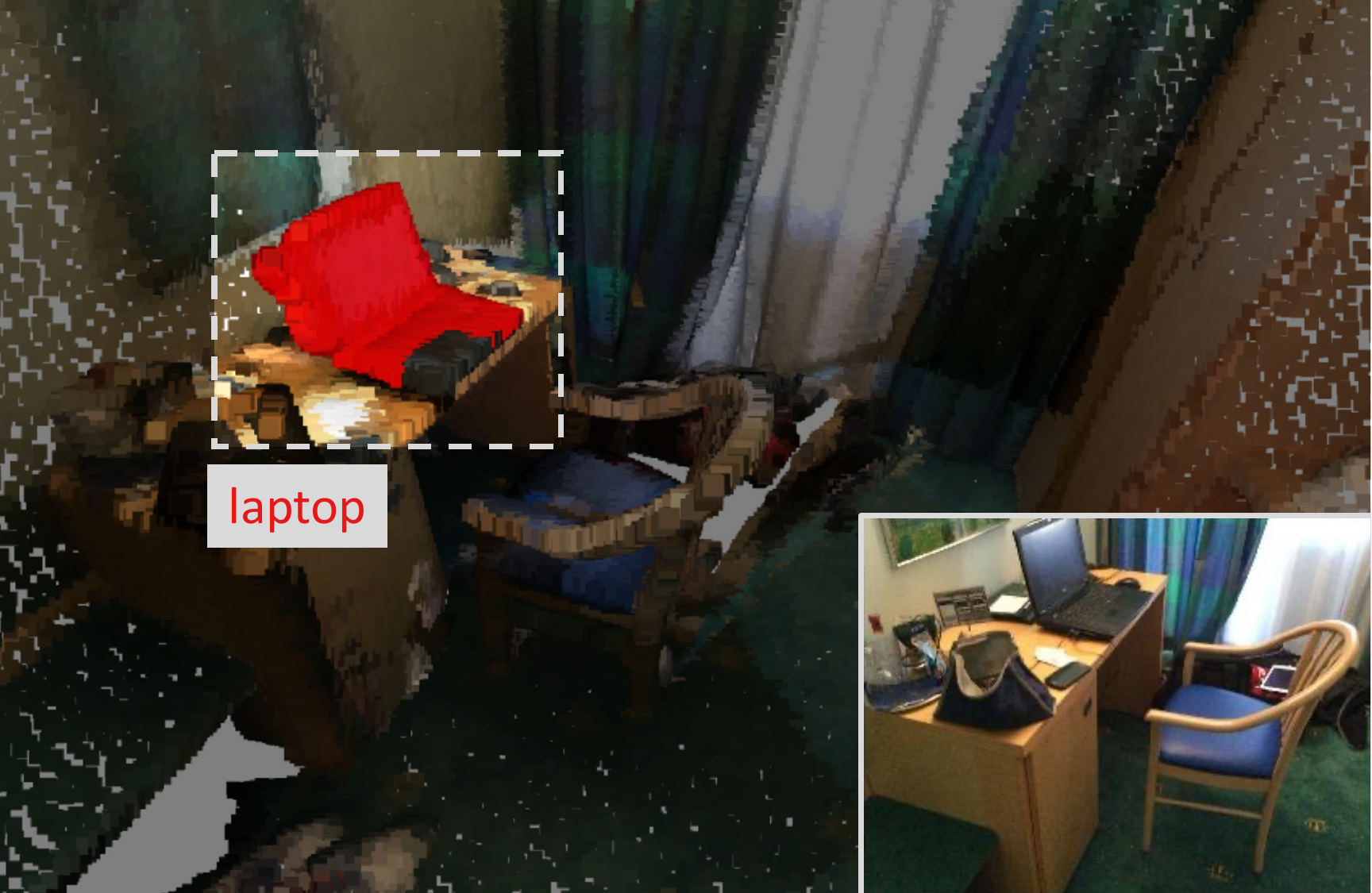}
    \caption{\textit{``Device to play game.''}}
  \end{subfigure}
  \begin{subfigure}{0.3233\linewidth}\includegraphics[width=1.0\linewidth]{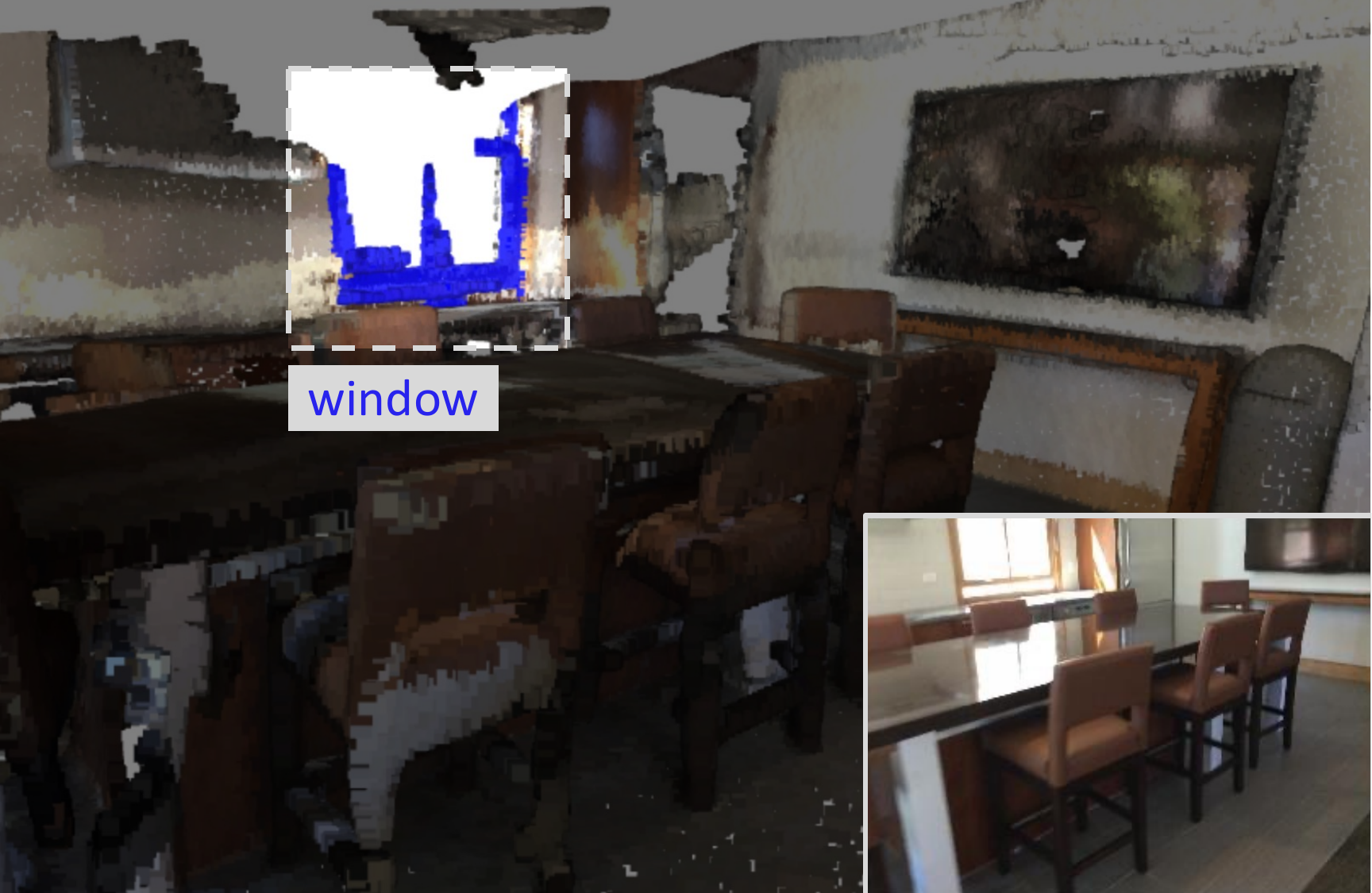}
    \caption{\textit{``I wanna see outside.''}}
  \end{subfigure}
  \caption{\textbf{Qualitative results when querying \ours with various language instructions.} \ours is highly generalizable and can segment corresponding instances with various language instructions, including but not limited to (a) \textit{visual questions}, (b) \textit{attributes description}, and (c) \textit{functional description}.}
  \label{fig:teaser}
\end{figure}
\vspace{-30pt}

\begin{abstract}
In this paper, we investigate Open-Vocabulary 3D Instance Segmentation (OV-3DIS) with free-form language instructions. Earlier works that rely on only annotated base categories for training suffer from limited generalization to unseen novel categories. Recent works mitigate poor generalizability to novel categories by generating class-agnostic masks or projecting generalized masks from 2D to 3D, but disregard semantic or geometry information, leading to sub-optimal performance. Instead, generating generalizable but semantic-related masks directly from 3D point clouds would result in superior outcomes. In this paper, we introduce Segment any 3D Object with LanguagE (\textbf{\ours}), which is a semantic and geometric-aware visual-language learning framework with strong generalizability by generating semantic-related masks directly from 3D point clouds. Specifically, we propose a multimodal fusion network to incorporate multimodal semantics in both backbone and decoder. In addition, to align the 3D segmentation model with various language instructions and enhance the mask quality, we introduce three types of multimodal associations as supervision. Our \ours outperforms previous methods by a large margin on ScanNetv2, ScanNet200, and Replica benchmarks, and the results are even closed to the fully-supervised counterpart despite the absence of class annotations in the training. Furthermore, extensive qualitative results demonstrate the versatility of our \ours to language instructions.

\end{abstract}

\section{Introduction}
\label{sec:intro}
3D instance segmentation %, 
which aims at detecting, segmenting and recognizing %the 
object instances in 3D scenes %, 
is one of the crucial tasks for 3D scene understanding. Effective and generalizable 3D instance segmentation has great potential in real-world applications, including but not limited to autonomous driving, augmented reality (AR), and virtual reality (VR). 
Owing to its significance, 3D instance segmentation has achieved remarkable success in the computer vision community~\cite{schult2022mask3d,vu2022softgroup,he2022pointinst3d}.
Previous 3D instance segmentation models mainly focus on the closed-set setting, where the training and testing stages share the same categories. However, novel and unseen categories with various shapes and semantic meaning are inevitable in real-world applications. Failure to segment such instances drastically narrows the scope of application.

In view of the strong limitations of closed-set setting, open-set 3D instance segmentation (OS-3DIS) that aims at detecting and segmenting unseen classes based on instructions
is introduced and investigated in the community. Most of the works~\cite{huang2023openins3d,ding2022pla,nguyen2023open3dis} leverage category names or descriptions as segmentation instructions, which is also termed as open-vocabulary 3D instance segmentation (OV-3DIS). The early approaches~\cite{ding2022pla, yang2023regionplc, ding2023lowis3d} split categories in each dataset into \textit{base} and \textit{novel} set. Only base categories are available in the training stage, but the model is expected to segment novel categories during inference. Due to the lack of novel classes during training, these methods easily overfit to the base categories, and thus yielding sub-optimal performance on novel categories. In addition, they suffer from severe performance degradation when they are evaluated on the data with different distributions. 
In this regards, recent works~\cite{takmaz2023openmask3d,huang2023openins3d,yan2024maskclustering,lu2023ovir} explore more generalizable OV-3DIS with the help of 2D foundation models~\cite{clip,oquab2023dinov2,detic}. Specifically,~\cite{takmaz2023openmask3d,huang2023openins3d} learn class-agnostic 3D masks from mask annotation and then project the point clouds to 2D images to obtain class labels from foundation models.~\cite{yan2024maskclustering,lu2023ovir} predict 2D instances with 2D open-vocabulary instance segmentation model~\cite{detic} and fuse them to obtain 3D predictions. However, class-agnostic masks and 2D projected masks ignore the semantic and geometry information in the mask generation, respectively, leading to the sub-optimal performance. 
In contrast, we directly predict semantic-related masks from 3D point clouds, yielding better and more generalizable 3D masks. 

In this paper, we propose \textbf{\ours}: \underline{S}egment any 3D \underline{O}bject with \underline{L}anguag\underline{E} to circumvent the above-mentioned issues for OV-3DIS. To realize generalizable open-set 3D instance segmentation, our \ours requires two main attributes: generating and classifying 3D masks directly from 3D point clouds, and responsive to free-form language instructions. The 3D segmentation network is required to be aligned with language instructions to directly segment and classify instances from point clouds. To this end, we build a multimodal fusion network with two main techniques: 1) Point-wise CLIP features obtained from pre-trained multimodal 2D semantic segmentation model~\cite{ghiasi2022scaling} are incorporated to the backbone to enhance the generality of the model; 2) Cross-modality decoder is introduced to integrate information from language-domain features, facilitating the effective fusion of multimodal knowledge. Furthermore, we improve the generalization ability across various scene and language instructions with a novel visual-language learning framework, training the 3D segmentation network with three types of multimodal associations: 1) Mask-visual association, 2) mask-caption association and 3) mask-entity association. These associations improve the language instruction alignment and enhance the 3D mask prediction with more abundant semantic information.

Equipped with a multimodal fusion network and three types of multimodal associations, our visual-language learning framework (\ours) outperforms previous works by a large margin on ScanNetv2~\cite{dai2017scannet}, ScanNet200~\cite{rozenberszki2022language} and Replica~\cite{straub2019replica} benchmarks. 
Furthermore, \ours can respond to free-form queries, including but not limited to questions, attributes description, and functional description (Fig.~\ref{fig:teaser} and Fig.~\ref{fig:qualitative}).
In summary, our contributions are as follows:

\begin{itemize}
\item We propose a visual-language learning framework for OV-3DIS, \ours. A multimodal fusion network is designed for \ours, which can directly predict semantic-related masks from 3D point clouds with multimodal information, leading to high-quality and generalizable segments.
\item We propose three types of multimodal associations to improve the alignment between 3D segmentation model with the language. The associations improve the mask quality and the response ability to language instructions.
\item \ours achieves state-of-the-art results on ScanNetv2, Scannet200 and Replica benchmarks, and the results are even close to the fully-supervised counterpart. In addition, extensive qualitative results demonstrate that \ours can respond to various language questions and instructions.
\end{itemize}

\section{Related Work}

\textbf{Closed-Set 3D Instance Segmentation.}  
3D instance segmentation aims at detecting, segmenting and recognizing the object instances in 3D scenes. Previous works~\cite{he2021dyco3d, he2022pointinst3d, ngo2023isbnet, schult2022mask3d, sun2023superpoint, vu2022softgroup, yi2019gspn, zhang2108pvt, hou20193d, yang2019learning, chen2021hierarchical, dong2022learning, jiang2020pointgroup, liu20223d, wu20223d} mainly consider the closed-set setting, where the training and testing categories are the same. These methods vary in feature extraction and decoding process. With the development of transformer models, mask prediction becomes a more efficient and effective way than traditional box detection decoding approaches. Mask3D~\cite{schult2022mask3d} samples a fixed number of points across the scene as queries, and then directly predicts the final masks with attention mechanism, achieving better results. 
However, closed-set methods lack the capability to handle the unseen categories regardless of the decoding approaches and thus hindering their application in the real world.

\vspace{2mm}
\noindent \textbf{Open-Vocabulary 2D Segmentation.} Owing to the recent success of large-scale vision-language models~\cite{alayrac2022flamingo, cherti2023reproducible, girdhar2023imagebind, jia2021scaling, clip, yu2022coca, yuan2021florence}, notable achievements have been made in open-vocabulary or zero-shot 2D segmentation~\cite{ding2022decoupling, ghiasi2022scaling, gu2021open, he2023open, kuo2022f, li2022languagedriven, liang2023open, ma2022open, rao2022denseclip, xu2022groupvit, xu2023open, zabari2021semantic, zhou2022extract, cho2023cat}. The common key idea is to leverage 2D mulitmodal foundation models~\cite{clip,jia2021scaling} for the transfer of image-level embeddings to the pixel-level downstream tasks. LSeg~\cite{li2022languagedriven}, OpenSeg~\cite{ghiasi2022scaling}, and OVSeg~\cite{liang2023open} align pixel-level or mask-level visual features to text features from foundation model for open-vocabulary semantic segmentation. Other works such as X-Decoder~\cite{zou2023generalized}, FreeSeg~\cite{qin2023freeseg} and SEEM~\cite{zou2024segment} suggest more unified-framework for open-vocabulary segmentation, include instance, panoptic, and referring segmentation.

\vspace{2mm}
\noindent \textbf{Open-Vocabulary 3D Scene Understanding.} 
The remarkable success achieved in open-vocabulary 2D segmentation (OV-2DS) has spurred several endeavors in open-vocabulary 3D segmentation. However, the techniques in OV-2DS cannot be directly transferred to the 3D domain due to the lack of 3D multimodal foundation model. Consequently, researchers propose to align 2D images and 3D point clouds and thus lifting 2D foundation models to 3D. For open-vocabulary 3D semantic segmentation,~\cite{chen2023open, ding2022pla, ha2022semantic, huang2023visual, jatavallabhula2023conceptfusion, peng2023openscene, shafiullah2022clip, shah2023lm} construct task-agnostic point-wise feature representations from 2D foundation models~\cite{clip}, and then use these features to query the open-vocabulary concepts within 3D scene. These works focus purely on transferring semantic information from 2D to 3D, limiting the application for instance-level recognition tasks. In this regard, open-vocabulary 3D instance segmentation (OV-3DIS)~\cite{ding2023lowis3d, huang2023openins3d, takmaz2023openmask3d, lu2023ovir, nguyen2023open3dis, yan2024maskclustering} is introduced to detect and segment instances of various categories in 3D scenes. PLA~\cite{ding2022pla} and its variants~\cite{yang2023regionplc, ding2023lowis3d} split the training categories into base and novel classes, and train the model only with base class annotation. OpenMask3D~\cite{takmaz2023openmask3d} and OpenIns3D~\cite{huang2023openins3d} learn class-agnostic 3D masks from mask annotations and then use the corresponding 2D images to obtain class labels from foundation models.
More recently, researchers also investigate direct lifting of 2D predictions from 2D instance segmentation model~\cite{detic} to 3D without training~\cite{yan2024maskclustering,lu2023ovir}. 
Previous works greatly prompt the improvement of OV-3DIS. However, the results are still far from satisfactory due to the poor semantic generalization ability and low-quality mask prediction.
Considering the limitations of previous work, we significantly improve OV-3DIS by designing a visual-language learning framework with a multimodal network and various multimodal associations.

\section{Method}

\noindent\textbf{Objective.}
The goal of open-vocabulary 3D instance segmentation (OV-3DIS) with free-form language instructions is defined as follows:
Given a 3D point cloud ${\mathbf{P}} \in \mathbb{R}^{M \times C}$, the corresponding 2D images $I$ and the instance-level 3D masks $\mathbf{m}$, we aim to train a 3D instance segmentation network {without ground-truth class annotations}. During inference, given a text prompt $q$, the trained 3D instance segmentation network must detect and segment corresponding instances.

\vspace{2mm}
\noindent{\textbf{Mask-Prediction Baseline.}} 
We build our framework on the transformer-based 3D instance segmentation model Mask3D~\cite{schult2022mask3d}, which treats the instance segmentation task as the mask prediction paradigm. Specifically, the transformer decoders with mask queries are used to segment instances. Given $N_q$ queries selected from the scene, cross attention is used to aggregate information from the point clouds {to instance queries}. After several decoder layers, $N_q$ queries become $N_q$ masks with corresponding semantic prediction. During training, Hungarian matching~\cite{kuhn1955hungarian} is adopted to match and train the model with ground truth labels and masks. At the inference stage, $N_q$ masks with correct semantic classification results are taken as the final outputs. Our \ours leverages the mask prediction paradigm with transformer-based architecture, where the model is only trained with masks without ground truth labels to achieve generalizable OV-3DIS.

\begin{figure}[tb]
  \centering
  \includegraphics[width=1.0\linewidth]{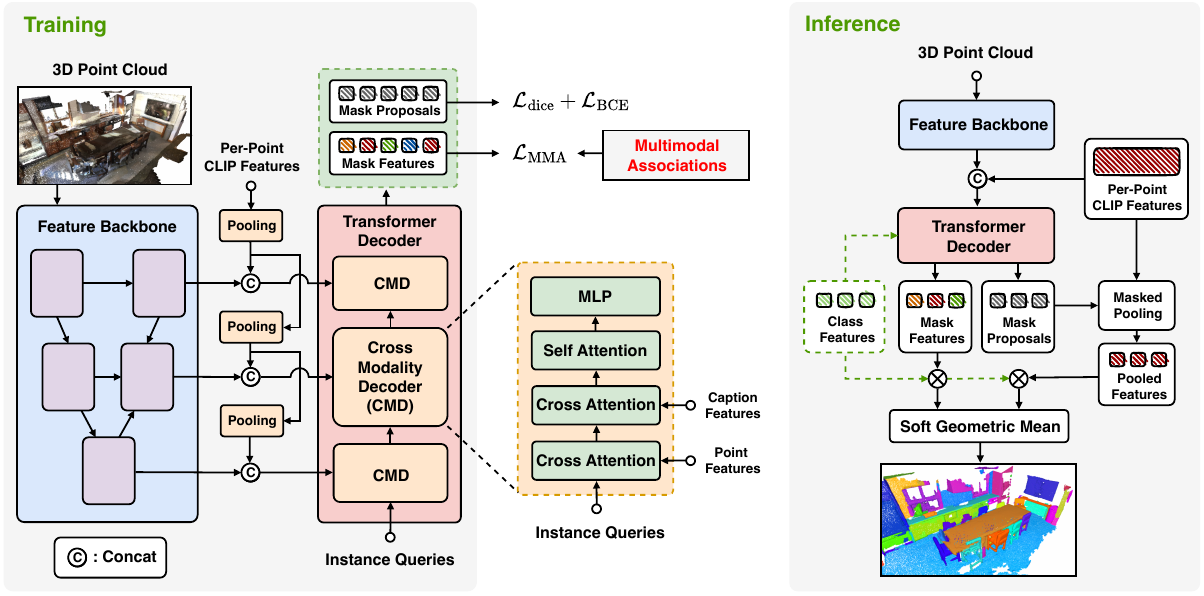}
  \caption{\textbf{Overall framework of \ours.} \ours is built on transformer-based instance segmentation model with multimodal adaptations. For model architecture, backbone features are integrated with per-point CLIP features and subsequently fed into the cross-modality decoder (CMD). CMD aggregates the point-wise features and textual features into the instance queries, finally segmenting the instances, which are supervised by multimodal associations. During inference, predicted mask features are combined with the per-point CLIP features, enhancing the open-vocabulary performance.}
  \label{fig:overall}
\vspace{-15pt}
\end{figure}

\vspace{2mm}
\noindent\textbf{Overview.} 
The overall architecture of \ours is illustrated in Fig.~\ref{fig:overall}. To realize open-vocabulary instance segmentation with free-form language instructions, we improve the transformer-based instance segmentation model with multimodal information: point-wise CLIP features in the backbone (Sec.~\ref{sec:ensemble}) and textual information in the decoder (Sec.~\ref{sec:cross-modal}). Furthermore, to achieve better generalization ability without ground truth class labels, we construct three types of multimodal associations on target instances: mask-visual association, mask-caption association and mask-entity association to train \ours. Equipped with the multimodal framework and associations, our \ours can effectively segment instances given various language prompts.

\subsection{Backbone Feature Ensemble}
\label{sec:ensemble}

Initializing the backbone with pre-trained model~\cite{vpt,zhao2023sct,zhao2023revisit} is an efficient and effective way to improve the performance on the downstream tasks, especially when the downstream data is not in abundance. For 3D open-set setting, leveraging 2D foundation model is crucial due to the limited 3D data. We thus follow \cite{peng2023openscene} to project pre-trained visual features of 2D images to 3D point clouds based on the camera pose. To maintain the fine-grained and generalizable features, we leverage OpenSeg~\cite{ghiasi2022scaling} as the 2D backbone. These features contain visual information in the CLIP~\cite{clip} feature space, which is aligned with textual information.

Since CLIP feature space mainly focuses on semantic information due to the image-level contrastive training, leveraging the projected features solely cannot achieve optimal performance on instance segmentation. To this end, we train a 3D instance segmentation backbone and combine its features $\mathbf{f}^{{b}} \in \mathbb{R}^{M \times D} $ with the projected 2D CLIP features $\mathbf{f}^{{p}} \in \mathbb{R}^{M \times C}$:
\begin{equation}
    \tilde{\mathbf{f}}^{{b}} = \mathrm{concat}(\mathbf{f}^{{p}}, \mathbf{f}^{{b}}) \in \mathbb{R}^{M \times (C+D)},
  % \label{eq:important}
\end{equation}
where $M$ denotes the number of points while $D$ and $C$ denote the feature dimension of 3D instance segmentation backbone and the projected 2D features, respectively. Note that features of different resolutions are extracted from the 3D backbone and respectively incorporated with the 2D CLIP features. As illustrated in Fig.~\ref{fig:overall}, the same pooling strategy with 3D backbone is adopted to CLIP features, aligning the resolution. Finally, incorporated point-wise features with multiple resolutions are fed into cross modality decoder.

\begin{figure}[tb]
  \centering
  \includegraphics[width=1.0\linewidth]{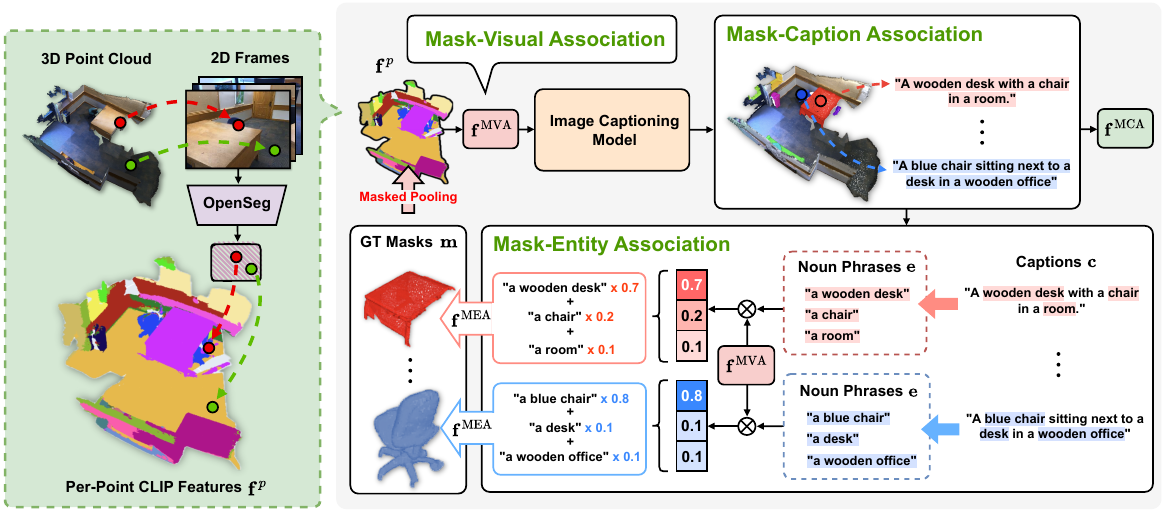}
  \caption{\textbf{{Three types of multimodal association instance.}} 
For each ground truth instance mask, we first pool the per-point CLIP features to obtain \textit{Mask-Visual Association} $\mathbf{f}^\text{MVA}$. Subsequently, $\mathbf{f}^\text{MVA}$ is fed into CLIP space captioning model to generate caption and corresponding textual feature $\mathbf{f}^\text{MCA}$ for each mask, termed as \textit{Mask-Caption Association}. Finally, noun phrases are extracted from mask caption and the embeddings of them are aggregated via multimodal attention to get \textit{Mask-Entity Association} $\mathbf{f}^\text{MEA}$. The three multimodal associations are used for supervising \ours to acquire the ability to segment 3D objects with free-form language instructions.
  }
  \label{fig:associations}
\vspace{-10pt}
\end{figure}

\subsection{Cross Modality Decoder (CMD)}
\label{sec:cross-modal}
Projected 2D CLIP features provide generalizable visual information but the language information is not explicitly integrated, limiting the responsive ability to language instructions. To circumvent this issue, we introduce Cross Modality Decoder (CMD) to incorporate textual information in the decoding process of our framework. Specifically, each CMD module contains three attention layers. Instance queries first extract visual information from the CLIP-combined backbone features $\tilde{\mathbf{f}}^{b}$. CLIP textual features are then projected to key and value in the second attention layer, incorporating the text domain knowledge. During the training, CLIP textual features are obtained from the caption features of each target mask, $\mathbf{f}^{\mathbf{MCA}} \in \mathbb{R}^{N_c \times C}$ (See Sec.~\ref{sec:mla} for details), whereas, during the inference, it can be the description of the query instance or other form of language instructions, such as visual questions or functional attributes. Finally, self-attention is applied to the instance queries to further improve the representation. By fusing the multimodal knowledge from CLIP with multi-level CMD as the decoder, \ours can respond to various language instructions with high-quality results.

\subsection{Vision-Language Learning}
\label{sec:mla}
 We do vision-language learning to enable our SOLE towards generalizable OV-3DIS. To respond effectively to various language instructions, we leverage multimodal information stemming from target mask annotations, to supervise the segmentation network. Specifically, three types of supervision in hierarchical granularity are proposed: 1) mask-visual association, 2) mask-caption association and 3) mask-entity association. 

\vspace{2mm}
\noindent \textbf{Mask-Visual Association (MVA).} 
Using the correspondence between 2D images and 3D point clouds, we can get the instance-level CLIP visual features $\mathbf{f}^\text{MVA} \in \mathbb{R}^{N_m \times C}$ by averaging the per-point CLIP features within the $N_m$ target instance masks $\mathbf{m} = [m_1, m_2, \ldots, m_{N_m}]$. The instance-level CLIP visual features can serve as the supervision to indirectly align the 3D segmentation model to CLIP textual space. In addition, as the intermediate representation between 3D point cloud and language, the mask-visual association is also the basis for the following two fine-grained associations.

\vspace{2mm}
\noindent \textbf{Mask-Caption Association (MCA).} 
Despite being in the CLIP feature space, mask-visual association is not an accurate and precision language supervision. Instead, directly supervising the model with language instructions would yield better results. Due to the strong generalization ability of CLIP~\cite{clip}, text generation from CLIP space is widely investigated in the community~\cite{tewel2022zerocap, mokady2021clipcap, li2023decap}. 
Since the instance-level CLIP visual features $\mathbf{f}^\text{MVA}$ in the mask-visual association is in CLIP visual space, we can feed them to the CLIP space caption generation model (DeCap~\cite{li2023decap}) to obtain the mask captions $\mathbf{c}=[c_1, c_2, ..., c_{N_m}]$. The mask captions are then fed into CLIP textual model to extract the mask-caption association $\mathbf{f}^\text{MCA}$. This association represents the language information for the instance masks, used in CMD to fuse textual information during the training.

\vspace{2mm}
\noindent \textbf{Mask-Entity Association (MEA).} 
Although mask-caption association can provide detailed language descriptions for both semantics and geometry, it may be ambiguous for specific categories. As shown in the example of Fig.~\ref{fig:associations}. The mask caption for a desk is ``A wooden desk with a chair in a room''. Such caption can lead to the confusion of the model between the chair and the desk, or misinterpretation of the two instances as a single one. It is therefore important to introduce a more fine-grained visual-language association for better semantic learning.

Since the objects are commonly the nouns in the caption, we can extract the entity-level descriptions for the nouns and match them with the instances. Specifically, as illustrated in Fig.~\ref{fig:associations}, we first extract all the noun phrases $\mathbf{e}_i$ for each mask caption $c_i$ and obtain the text feature of each noun phrase from CLIP text encoder $\mathcal{T}$ as below:
\begin{equation}
    \mathcal{E}(c_i) = \mathbf{e}_i = [e_1, e_2, \dots, e_{N_e^i}], \quad \mathbf{f}^{\mathbf{e}}_i = \mathcal{T}(\mathbf{e}_i) \in \mathbb{R}^{{N_e^i} \times C},
\end{equation}
where $\mathcal{E}(\cdot)$ denotes the NLP tool to extract noun phrases and ${N_e^i}$ denotes the number of nouns obtained from mask caption $c_i$. The entities can be matched to the mask in either a hard or soft manner. Intuitively, the most similar entity can be viewed as the mask label. However, there are two main issues with such a hard matching. First, the generated caption and the similarity results may not be accurate, leading to wrong supervision. Second, although the entity is correct, hard matching ignores the geometry information in the context and thus impairing the responsive ability to language instructions. To this end, we propose a soft matching to get mask-entity association by multimodal attention. Specifically, the aggregated entity feature for the $i$-th mask $\mathbf{f}^\text{MEA}_i$ is obtained based on the attention map $A_{c,e}$ between mask feature and entity features:
\begin{equation}
    \mathbf{f}^\text{MEA}_i = A_{c,e} \cdot \mathbf{f}^{\mathbf{e}}_i  = \sum_k^{N_e^i} \frac{\exp{\left(\mathbf{f}^\text{MVA}_i \cdot \mathbf{f}^{e_k}_i\right) }}{\sum_j^{N_e^i} \exp{\left(\mathbf{f}^\text{MVA}_i \cdot \mathbf{f}^{e_j}_i\right) } } \cdot \mathbf{f}^{e_k}_i,
\end{equation}
where $\mathbf{f}^\text{MVA}_i$ denotes the mask-visual association feature for $i$-th mask, and $\mathbf{f}^{e_k}_i$ is the CLIP textual feature for $k$-th entity in the $i$-th mask caption. With the aggregated entity feature, the 3D mask can be aligned with a specific instance category.

\subsection{Training and Inference}
\label{sec:training}

\noindent\textbf{Training.}
The three types of multimodal associations are effective supervision to learn a generalizable 3D instance segmentation model. We follow the mask prediction paradigm to train the segmentation model, which matches the ground truth instances with the predicted masks via Hungarian matching~\cite{kuhn1955hungarian}. Specifically, the matching cost between $i$-th predicted mask and $j$-th ground truth instance is calculated as:
\begin{equation}
    \begin{aligned}
    \mathcal{C}(i, j) = &-\lambda_{\mathrm{MMA}}\left( p\left(\mathbf{f}^{\mathbf{m}}_i \cdot \mathbf{f}^\text{MVA}_j \right) + p\left(\mathbf{f}^{\mathbf{m}}_i \cdot \mathbf{f}^\text{MCA}_j \right)  + p\left(\mathbf{f}^{\mathbf{m}}_i \cdot \mathbf{f}^\text{MEA}_j \right)  \right) \\
    &+ \lambda_{\mathrm{dice}}\mathcal{L}_{\mathrm{dice}}(i, j) + \lambda_{\mathrm{BCE}}\mathcal{L}_{\mathrm{BCE}}(i, j),
    \end{aligned}
\end{equation}
where $p(\cdot, \cdot)$ denotes the softmax probability between the predicted instance and the ground truth. %one. 
After matching the masks and ground truth instances, the model is trained with the combination of mask and semantic loss. 
Specifically, all three types of associations are used to semantically supervise the model. For each association, we follow~\cite{zhou2023zegclip} to use the combination of focal loss~\cite{lin2017focal} and dice loss, which can ensure the segmentation result for each class is independently generated. The semantic multimodal association loss $\mathcal{L}_{\mathrm{MMA}}^j$ for $j$-th ground truth mask is:
\begin{equation}
    \mathcal{L}_{\mathrm{MMA}}^j = \sum_a \left( \mathcal{L}_{\mathrm{focal}}(\hat{p}_{\sigma(j)}^a, y_j^a) + \mathcal{L}_{\mathrm{dice}} (\hat{p}_{\sigma(j)}^a, y_j^a) \right),
\end{equation}
where $a \in \{\mathrm{MVA}, \mathrm{MCA}, \mathrm{MEA}\}$ denotes three types of associations and $y_j^a$ is the binary label for matching. $\hat{p}_{\sigma(j)}^a = \mathrm{sigmoid}(\mathbf{f}^{\mathbf{m}}_{\sigma(j)} \cdot \mathbf{f}^{a}_j)$ is the semantic probability between the prediction with the association $a$.
The overall training loss is the combination of mask loss and semantic loss:
\small
\begin{equation}
    \mathcal{L} = \frac{1}{N_m} \sum_j^{N_m} \left( 
    \lambda_{\mathrm{MMA}} \mathcal{L}_{\mathrm{MMA}}^j + 
    \lambda_{\mathrm{dice}}\mathcal{L}_{\mathrm{dice}}(\hat{m}_{\sigma(j)}, m_j) + \lambda_{\mathrm{BCE}}\mathcal{L}_{\mathrm{BCE}}(\hat{m}_{\sigma(j)}, m_j)  \right),
\end{equation}
\normalsize
where $\hat{m}_{\sigma(j)}$ denotes matched predicted mask with $j$-th target mask.

\vspace{2mm}
\noindent\textbf{Inference.}
During inference, we combine the visual feature from CLIP with the predicted mask feature to achieve better generalization ability. Specifically, after obtaining the 3D masks, per-point CLIP features are pooled within the mask. The pooled CLIP feature and mask feature are then fed into the classifier to obtain the respective classification probability $p(\mathbf{f}^m)$ and $p(\mathbf{f}^p)$, and the final probability is yielded by soft geometric mean between them:
\begin{equation}\label{eq:soft_geometric_mean}
    p = \max\left(p(\mathbf{f}^m),p(\mathbf{f}^p)\right)^\tau \cdot \min\left(p(\mathbf{f}^m),p(\mathbf{f}^p)\right)^{1-\tau},
\end{equation}
where $\tau$ is the exponent to increase confidence, which we set to $0.667$ in this paper.
For benchmark evaluation, we use CLIP textual features of all category names as the classifier. For responding to other language instructions, we use the CLIP textual feature of corresponding language instruction as binary classifier.

\section{Experiments}

\begin{table}[t]
    \centering
    \caption{\textbf{The comparison of closed-set 3D instance segmentation setting on ScanNetv2~\cite{dai2017scannet}.} 
    \ours is compared with class-split methods, mask-training methods and the full-supervised counterpart (upper bound). \ours outperforms all the OV-DIS methods and achieves competitive results with the fully-supervised model.
    }
    \label{tab:scannet}
    \renewcommand{\arraystretch}{1.1}
    \small
    \begin{tabular}{>{\arraybackslash}p{3.5cm}|
    >{\centering\arraybackslash}p{1.1cm}
    >{\centering\arraybackslash}p{1.1cm}
    >{\centering\arraybackslash}p{1.1cm}
    >{\centering\arraybackslash}p{1.1cm}|
    >{\centering\arraybackslash}p{1.5cm}}
    \toprule
    Method & B/N & AP & AP$_{50}$ & AP$_{25}$ & voxel size \\
    \midrule
    PLA~\cite{ding2022pla} & 10/7 & - & 21.9 & -  & 2cm \\
    RegionPLC~\cite{yang2023regionplc} & 10/7 & - & 32.3 & -  & 2cm \\
    Lowis3D~\cite{ding2023lowis3d}  & 10/7 & - & 31.2 & -  & 2cm \\
    OpenIns3D~\cite{huang2023openins3d}  & -/7 & - & 27.9 & 42.6  & 2cm \\
    \hline
    \ours \textit{w 4cm voxel size}  & -/7 & 31.6 & 58.5 & 72.5  & 4cm \\
    \ours \textit{w/o text sup}  & -/7 & 41.1 & 57.1 & 65.9  & 2cm \\
    \rowcolor[gray]{0.8} \textbf{\ours (\textit{ours})}  & -/7 & \textbf{52.3} & \textbf{72.4}  & \textbf{81.7}  & 2cm \\ 
    
    \midrule
    PLA~\cite{ding2022pla}  & 8/9 & - & 25.1 & -  & 2cm \\
    RegionPLC~\cite{yang2023regionplc}  & 8/9 & - & 32.2 & -  & 2cm \\
    Lowis3D~\cite{ding2023lowis3d}  & 8/9 & - & 38.1 & -  & 2cm \\
    OpenIns3D~\cite{huang2023openins3d}  & -/9 & - & 19.5 & 27.9  & 2cm \\
    \hline
    \ours \textit{w 4cm voxel size}  & -/9 & 31.9 & 57.5 & 73.6  & 4cm \\
    \ours \textit{w/o text sup}  & -/9 & 42.9 & 59.6 & 70.7  & 2cm \\
    \rowcolor[gray]{0.8} \textbf{\ours (\textit{ours})}  & -/9 & \textbf{50.4} & \textbf{68.3}  & \textbf{75.2}  & 2cm \\ 
    \midrule
    OpenIns3D~\cite{huang2023openins3d}  & -/17 & - & 28.7 & 38.9  & 2cm \\
    \hline
    \ours \textit{w 4cm voxel size}  & -/17 & 30.8 & 52.5 & 70.9  & 4cm \\
    \ours \textit{w/o text sup}  & -/17 & 35.0 & 50.2 & 60.2  & 2cm \\
    \rowcolor[gray]{0.8} \textbf{\ours (\textit{ours})}  & -/17 & \textbf{44.4} & \textbf{62.2} & \textbf{71.4}  & 2cm \\
    \midrule
    Mask3D~\cite{schult2022mask3d} (\textit{fully sup}) & 17/- & 55.2 & 73.7 & 83.5  & 2cm \\
  \bottomrule
    \end{tabular}
    \vspace{-15pt}
\end{table}

\begin{table}[t]
  \caption{\textbf{The comparison of closed-set 3D instance segmentation setting on ScanNet200~\cite{rozenberszki2022language}.} 
  \ours is compared with OpenMask3D~\cite{takmaz2023openmask3d} on the overall segmentation performance and on each subset. \ours significantly outperforms OpenMask3D on five out of the six evaluation metrics.
  }
  \vspace{-.1in}
  \label{tab:scannet200}
  \centering
  \small
  \renewcommand{\arraystretch}{1.1}
  \begin{tabularx}{.85\textwidth}
  { 
  >{\hsize=.8\hsize\linewidth=\hsize}X
  >{\centering\arraybackslash\hsize=.3\hsize\linewidth=\hsize}X 
  >{\centering\arraybackslash\hsize=.3\hsize\linewidth=\hsize}X 
  >{\centering\arraybackslash\hsize=.3\hsize\linewidth=\hsize}X 
  >{\centering\arraybackslash\hsize=.3\hsize\linewidth=\hsize}X 
  >{\centering\arraybackslash\hsize=.3\hsize\linewidth=\hsize}X
  >{\centering\arraybackslash\hsize=.3\hsize\linewidth=\hsize}X}
    \toprule
    {Method} & {AP} & {AP$_{50}$} & {AP$_{25}$} & {AP$_{head}$} & {AP$_{com}$} & {AP$_{tail}$}\\
    \midrule
    {OpenMask3D~\cite{takmaz2023openmask3d}} & {15.4} & {19.9} &  {23.1} & {17.1} & {14.1} & {\textbf{14.9}} \\
    \rowcolor[gray]{0.8} \textbf{\ours (\textit{ours})} & {\textbf{20.1}} \textcolor{lightgreen}{\tiny{(+4.7)}} & {\textbf{28.1}} \textcolor{lightgreen}{\tiny{(+8.2)}} & {\textbf{33.6}} \textcolor{lightgreen}{\tiny{(+10.5)}} & {\textbf{27.5}} \textcolor{lightgreen}{\tiny{(+10.4)}} & {\textbf{17.6}} \textcolor{lightgreen}{\tiny{(+3.5)}} & {14.1} \textcolor{lightred}{\tiny{(-0.8)}} \\
    \midrule
    {Mask3D~\cite{schult2022mask3d}} & {26.9} & {36.2} &  {41.4} & {39.8} & {21.7} & {17.9} \\
  \bottomrule
  \end{tabularx}
  \vspace{-.1in}
\end{table}

\subsection{Experimental Setting}
\label{sec:experiment-setting}

\noindent\textbf{Datasets.}
We evaluate \ours on the popular scene understanding datasets: ScanNetv2~\cite{dai2017scannet}, ScanNet200~\cite{rozenberszki2022language} and Replica~\cite{straub2019replica} in both closed-set and open-set 3D instance segmentation tasks. 
ScanNetv2~\cite{dai2017scannet} is a popular indoor point cloud dataset with 18 instance classes, %while
where ``other furniture'' class is disregarded due to its ambiguity. ScanNet200~\cite{rozenberszki2022language} is a fine-grained annotated version of ScanNetv2 %, containing
that contains 200 classes of head (66 categories), common (68 categories) and tail (66 categories) subsets.
For ScanNetv2 and ScanNet200, we evaluate the closed-set setting and the hierarchical open-set setting.
Replica~\cite{straub2019replica} is a high-quality synthetic dataset annotated with 48 instance categories. 
Following \cite{takmaz2023openmask3d}, we evaluate on eight scenes in Replica for open-set instance segmentation, including \{office0, office1, office2, office3, office4, room0, room1 and room2.\}

\vspace{2mm}
\noindent\textbf{Implementation Details.}
Following the Mask3D~\cite{schult2022mask3d}, we adopt MinkowskiUNet~\cite{choy20194d} as backbone. The feature backbone extracts point features in 5 scales, while 4 layers of transformer decoder iteratively refine the instance queries. Our model is trained for 600 epochs with AdamW~\cite{loshchilov2017decoupled} optimizer. The learning rate is set to $1\times 10^{-4}$ with cyclical decay. In training, we set $\lambda_{MMA}$ = 20.0, $\lambda_{dice}$ = 2.0 and $\lambda_{BCE}$ = 5.0 as the loss weight. 

\vspace{2mm}
\noindent\textbf{Baselines.}
We compare \ours mainly with two streams of existing works on OV-3DIS: class-split methods~\cite{ding2022pla, yang2023regionplc, ding2023lowis3d} and mask-training methods~\cite{takmaz2023openmask3d,huang2023openins3d}.
Class-split methods~\cite{ding2022pla, yang2023regionplc, ding2023lowis3d} split the training categories into base and novel categories. All the mask annotations and base category labels are used to train the model. When compared with these methods, we only train our model on the mask annotation and compare with them on the split novel categories. Mask-training methods~\cite{takmaz2023openmask3d,huang2023openins3d} train class-agnostic mask generator with mask annotation and get the semantic prediction with 2D foundation models. The setting of mask-training methods is similar to ours, and we directly compare with them on all the categories.

\vspace{2mm}
\noindent\textbf{Evaluation Metric.}
Average precision (AP) of different IoU thresholds is adopted as the evaluation metric, including AP under 25\%, 50\% IoU and the average AP from 50\% to 95\% IoU.

\subsection{Comparison with Previous Methods}
\label{sec:results}

\noindent \textbf{Closed-Set 3D Instance Segmentation.} 
We compare our \ours with both class-split methods~\cite{ding2022pla,yang2023regionplc,ding2023lowis3d} and mask-training methods~\cite{takmaz2023openmask3d,huang2023openins3d} on the closed-set 3D instance segmentation setting. When compared with class-split methods, we evaluate on the novel categories. 
From the comparison results in Tab.~\ref{tab:scannet}, we can make the following observations.
\textbf{First}, \ours significantly outperforms the class-split methods by a large margin, even without using the base class labels. \textbf{Second}, although OpenIns3D~\cite{huang2023openins3d} leverages the same mask annotation with our \ours, we significantly surpasses it by 33.5\% and 32.5\% on AP$_{50}$ and AP$_{25}$, respectively. \textbf{Third}, our \ours can even achieve competitive performance with the fully-supervised counterpart (44.4\% \textit{v.s.} 55.2\% in AP) despite not using the class labels. Finally, we provide two variants of \ours to further verify our effectiveness. \ours \textit{w 4cm voxel size} leverages 4cm voxel size instead of 2cm as in previous works. Smaller voxel size can save the memory requirements and speed up the model with the loss of precision. Despite using a small voxel size, \ours \textit{w 4cm voxel size} can still outperform previous works by a large margin. Furthermore, we verify that the effectiveness of our framework is not limited to the caption model and NLP tools by conducting experiments without any additional textual information, \ie \ours \textit{w/o text sup}. In this experiment, mask-caption association and mask-entity association is removed since the caption is not available. The model can still achieve the state-of-the-art performance despite only trained with mask-visual association. Additionally, we compare \ours with OpenMask3D~\cite{takmaz2023openmask3d} on ScanNet200~\cite{rozenberszki2022language} in Tab.~\ref{tab:scannet200}. The two methods are evaluated on the overall segmentation performance and the performance on each of the three subsets. \ours outperforms OpenMask3D~\cite{takmaz2023openmask3d} on five out of six metrics and achieves comparable performance on the tail classes. The results on ScanNet200~\cite{rozenberszki2022language} further demonstrate the effectiveness of our framework.

\begin{table}[tb]
  \caption{
\textbf{The comparison of hierarchical open-set 3D instance segmentation setting on ScanNetv2~\cite{dai2017scannet}$\rightarrow$ScanNet200~\cite{rozenberszki2022language}.} 
  \ours is compared with OpenMask3D~\cite{takmaz2023openmask3d} on both base and novel classes and achieves the best results.
  }
  \vspace{-.1in}
  \label{tab:scannet20-scannet200}
  \centering
  \renewcommand{\arraystretch}{1.1}
  \begin{tabularx}{\textwidth}
  { 
  >{\hsize=.6\hsize\linewidth=\hsize}X |
  >{\centering\arraybackslash\hsize=.2\hsize\linewidth=\hsize}X 
  >{\centering\arraybackslash\hsize=.2\hsize\linewidth=\hsize}X 
  >{\centering\arraybackslash\hsize=.2\hsize\linewidth=\hsize}X |
  >{\centering\arraybackslash\hsize=.2\hsize\linewidth=\hsize}X
  >{\centering\arraybackslash\hsize=.2\hsize\linewidth=\hsize}X 
  >{\centering\arraybackslash\hsize=.2\hsize\linewidth=\hsize}X |
  >{\centering\arraybackslash\hsize=.2\hsize\linewidth=\hsize}X  
  >{\centering\arraybackslash\hsize=.2\hsize\linewidth=\hsize}X}
    \toprule
     \multirow{2}{*}{Method} & \multicolumn{3}{c|}{{\textit{Novel Classes}}} & \multicolumn{3}{c|}{{\textit{Base Classes}}} & \multicolumn{2}{c}{{\textit{All Classes}}}\\
     &  {AP} & {AP$_{50}$} & {AP$_{25}$} & {AP} & {AP$_{50}$} & {AP$_{25}$} & {AP} & {AP$_\text{tail}$} \\
    \midrule
    {OpenMask3D~\cite{takmaz2023openmask3d}} & {11.9} & {15.2} & {17.8} & {14.3} & {18.3} & {21.2} & {12.6} & {11.5} \\
    \rowcolor[gray]{0.8} {\textbf{\ours (\textit{ours})}}  & {\textbf{19.1}} \textcolor{lightgreen}{\tiny{(+8.8)}} & {\textbf{26.2}} \textcolor{lightgreen}{\tiny{(+11.0)}} & {\textbf{30.7}} \textcolor{lightgreen}{\tiny{(+12.9)}} & {\textbf{17.4}} \textcolor{lightgreen}{\tiny{(+3.1)}} & {\textbf{26.2}} \textcolor{lightgreen}{\tiny{(+7.9)}} & {\textbf{32.1}} \textcolor{lightgreen}{\tiny{(+10.9)}} & {\textbf{18.7}} \textcolor{lightgreen}{\tiny{(+6.1)}} & {\textbf{12.5}} \textcolor{lightgreen}{\tiny{(+1.0)}} \\
  \bottomrule
  \end{tabularx}
  \vspace{-3pt}
\end{table}

\begin{table}[tb]
\caption{
\textbf{The comparison of open-set 3D instance segmentation setting on ScanNet200~\cite{rozenberszki2022language}$\rightarrow$Replica~\cite{straub2019replica}.} 
\ours outperforms OpenMask3D~\cite{takmaz2023openmask3d} on all the evaluation metrics.
}
  \vspace{-.1in}
  \label{tab:scannet200-replica}
  \centering
  \small
  \begin{tabularx}{0.9\textwidth}
  { 
  >{\hsize=.5\hsize\linewidth=\hsize}X 
  >{\hsize=.5\hsize\linewidth=\hsize}X
  >{\hsize=.3\hsize\linewidth=\hsize}X 
  >{\hsize=.3\hsize\linewidth=\hsize}X 
  >{\hsize=.3\hsize\linewidth=\hsize}X}
    \toprule
    {Method} & {Mask Training} & {AP} & {AP$_{50}$} & {AP$_{25}$} \\
    \midrule
    {OpenMask3D~\cite{takmaz2023openmask3d}} & {ScanNet200~\cite{rozenberszki2022language}} & {13.1} & {18.4} & {24.2} \\
    \rowcolor[gray]{0.8} {\textbf{\ours (\textit{ours})}} & {ScanNet200~\cite{rozenberszki2022language}} & {\textbf{24.7}} \textcolor{lightgreen}{\tiny{(+11.6)}} & {\textbf{31.8}} \textcolor{lightgreen}{\tiny{(+13.4)}} & {\textbf{40.3}} \textcolor{lightgreen}{\tiny{(+16.1)}} \\
  \bottomrule
  \end{tabularx}
  \vspace{-8pt}
\end{table}

\vspace{2mm}
\noindent\textbf{Hierarchical and Cross-Domain Open-Set 3DIS.}
To evaluate the generalization capability of our work, we compare our \ours with OpenMask3D~\cite{takmaz2023openmask3d} in open-set setting, using Scannet200~\cite{rozenberszki2022language} and Replica~\cite{straub2019replica} datasets. 
For ScanNet200, both models are trained with mask annotations in ScanNetv2~\cite{dai2017scannet}. 
Following \cite{takmaz2023openmask3d}, 53 classes that are semantically close to the ScanNet, are grouped as “Base”.
The remaining 147 classes are grouped as “Novel”.
Both in-distribution (``base'') and out-of-distribution (``novel'') classes are reported in Tab.~\ref{tab:scannet20-scannet200}. Our \ours outperforms OpenMask3D~\cite{takmaz2023openmask3d} by a large margin on both base and novel classes.
Furthermore, to verify the generalization ability of \ours when both domain shift and category shift exist, we compare our framework with OpenMask3D on the synthetic Replica benchmark~\cite{straub2019replica}. Models are trained on the annotated masks on ScanNet200.
As shown in Tab.~\ref{tab:scannet200-replica}, our method further shows superior robustness on more out-of-distribution data from Replica, achieving +11.6\% improvement in AP score compared to OpenMask3D.

\subsection{Ablation Studies and Analysis}
\label{sec:abl}
In this section, we conduct several ablation studies to validate our design choices. All of the studies are evaluated on ScanNetv2~\cite{dai2017scannet} dataset.

\vspace{2mm}
\noindent\textbf{Multimodel Fusion Network.}
In Tab.~\ref{tab:fusion-network}, we conduct component analysis on multimodal fusion network, validating the effectiveness of backbone feature ensemble and Cross-Modality Decoder (CMD). As for the backbone feature ensemble, leveraging projected 2D CLIP features $\mathbf{f}^p$ (first row) as only backbone can have better semantic information but lack the 3D geometry detection ability, leading to poor semantic recognition ability. In contrast, solely using 3D instance backbone feature $\mathbf{f}^b$ (second row) cannot inherit the generalizable semantic information, resulting in sub-optimal performance. Combining the two features (third row) can make full use of generalized semantic information while learning good geometry detection ability from 3D masks, yielding optimal results.
Additionally, Cross Modality Decoder (CMD) can further enhance the ability to understand language instructions, improving AP by 1.6\%.

\begin{table}[tb]
  \caption{\textbf{Component analysis on multimodal fusion network.} 3D instance backbone feature $\mathbf{f}^b$, projected 2D backbone feature $\mathbf{f}^p$, and Cross Modality Decoder (CMD) are investigated separately.
  }
  \vspace{-.1in}
  \small
  \label{tab:fusion-network}
  \centering
  \renewcommand{\arraystretch}{1.1}
  \begin{tabularx}{0.8\textwidth}
  { 
  >{\centering\arraybackslash\hsize=.2\hsize\linewidth=\hsize}X |
   >{\centering\arraybackslash\hsize=.2\hsize\linewidth=\hsize}X 
  >{\centering\arraybackslash\hsize=.2\hsize\linewidth=\hsize}X 
  >{\centering\arraybackslash\hsize=.2\hsize\linewidth=\hsize}X |
  >{\centering\arraybackslash\hsize=.3\hsize\linewidth=\hsize}X 
  >{\centering\arraybackslash\hsize=.3\hsize\linewidth=\hsize}X 
  >{\centering\arraybackslash\hsize=.3\hsize\linewidth=\hsize}X |
  >{\centering\arraybackslash\hsize=.4\hsize\linewidth=\hsize}X }
    \toprule
    No. & $\mathbf{f}^p$ & $\mathbf{f}^b$ & CMD & {AP} & {AP$_{50}$} & {AP$_{25}$} & {voxel size} \\
    \midrule
    1 & \ding{51} &  & \ding{51} & {18.7} & {36.4} & {58.1} & {4cm} \\
    2 &  & \ding{51} & \ding{51} & {25.4} & {47.0} & {66.0} & {4cm} \\
    3 & \ding{51} & \ding{51} & \ding{51} & {\textbf{30.8}} & {\textbf{52.5}} & {\textbf{70.9}} & {4cm} \\
     \midrule
    4 & \ding{51} & \ding{51} & {} & {42.8} & {60.5} & {68.9} & {2cm} \\
     5 & \ding{51} & \ding{51} & \ding{51} & {\textbf{44.4}} & {\textbf{62.2}} & {\textbf{71.4}} & {2cm} \\
    \bottomrule
  \end{tabularx}
  \vspace{-4pt}
\end{table}

\begin{table}[tb]
  \caption{\textbf{Component Analysis on multimodal associations.} Mask-visual association $\mathbf{f}^\text{MVA}$, mask-caption association $\mathbf{f}^\text{MCA}$ and mask-entity association $\mathbf{f}^\text{MEA}$ are studied on ScanNetv2~\cite{dai2017scannet}.
  }
  \vspace{-.1in}
  \label{tab:multimodal-associations}
  \centering
  \renewcommand{\arraystretch}{1.1}
  \small
  \begin{tabularx}{0.8\textwidth}
  { 
  >{\centering\arraybackslash\hsize=.2\hsize\linewidth=\hsize}X |
   >{\centering\arraybackslash\hsize=.2\hsize\linewidth=\hsize}X 
  >{\centering\arraybackslash\hsize=.2\hsize\linewidth=\hsize}X 
  >{\centering\arraybackslash\hsize=.2\hsize\linewidth=\hsize}X |
  >{\centering\arraybackslash\hsize=.3\hsize\linewidth=\hsize}X 
  >{\centering\arraybackslash\hsize=.3\hsize\linewidth=\hsize}X 
  >{\centering\arraybackslash\hsize=.3\hsize\linewidth=\hsize}X |
  >{\centering\arraybackslash\hsize=.4\hsize\linewidth=\hsize}X }
    \toprule
    No. & $\mathbf{f}^\text{MVA}$ & $\mathbf{f}^\text{MCA}$ & $\mathbf{f}^\text{MEA}$ & {AP} & {AP$_{50}$} & {AP$_{25}$} & {voxel size} \\
    \midrule
    1 & \ding{51} &  &  & {24.5} & {42.0} & {56.0} & {4cm} \\
    2 &  & \ding{51} &  & {30.4} & {53.0} & {68.7} & {4cm} \\
    3 &  &  & \ding{51} & {\textbf{32.1}} & {\textbf{53.8}} & {70.0} & {4cm} \\
    4 & \ding{51} & \ding{51} &  & {29.1} & {50.9} & {66.8} & {4cm} \\
    5 & & \ding{51} & \ding{51} & {30.3} & {53.7} & {70.4} & {4cm} \\
    6 & \ding{51} & \ding{51} & \ding{51} & {30.8} & {52.5} & {\textbf{70.9}} & {4cm}  \\
    \bottomrule
  \end{tabularx}
  \vspace{-8pt}
\end{table}

\begin{figure}[tb]
\captionsetup[subfigure]{labelformat=empty}
  \centering
  \begin{subfigure}{0.323\linewidth}\includegraphics[width=1.0\linewidth]{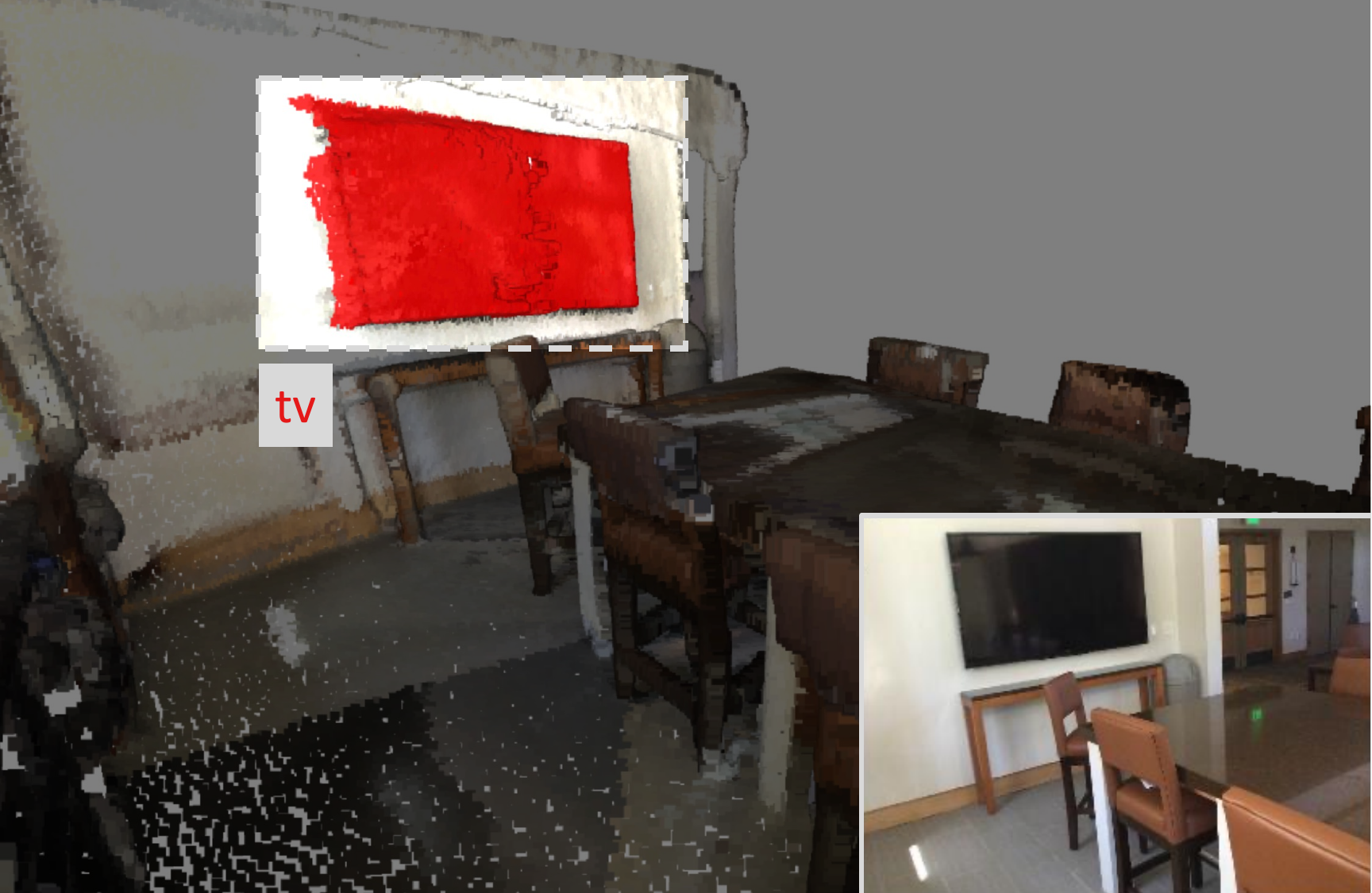}
    \caption{\textit{``I want to watch movie.''}}
  \end{subfigure}
  \begin{subfigure}{0.323\linewidth}\includegraphics[width=1.0\linewidth]{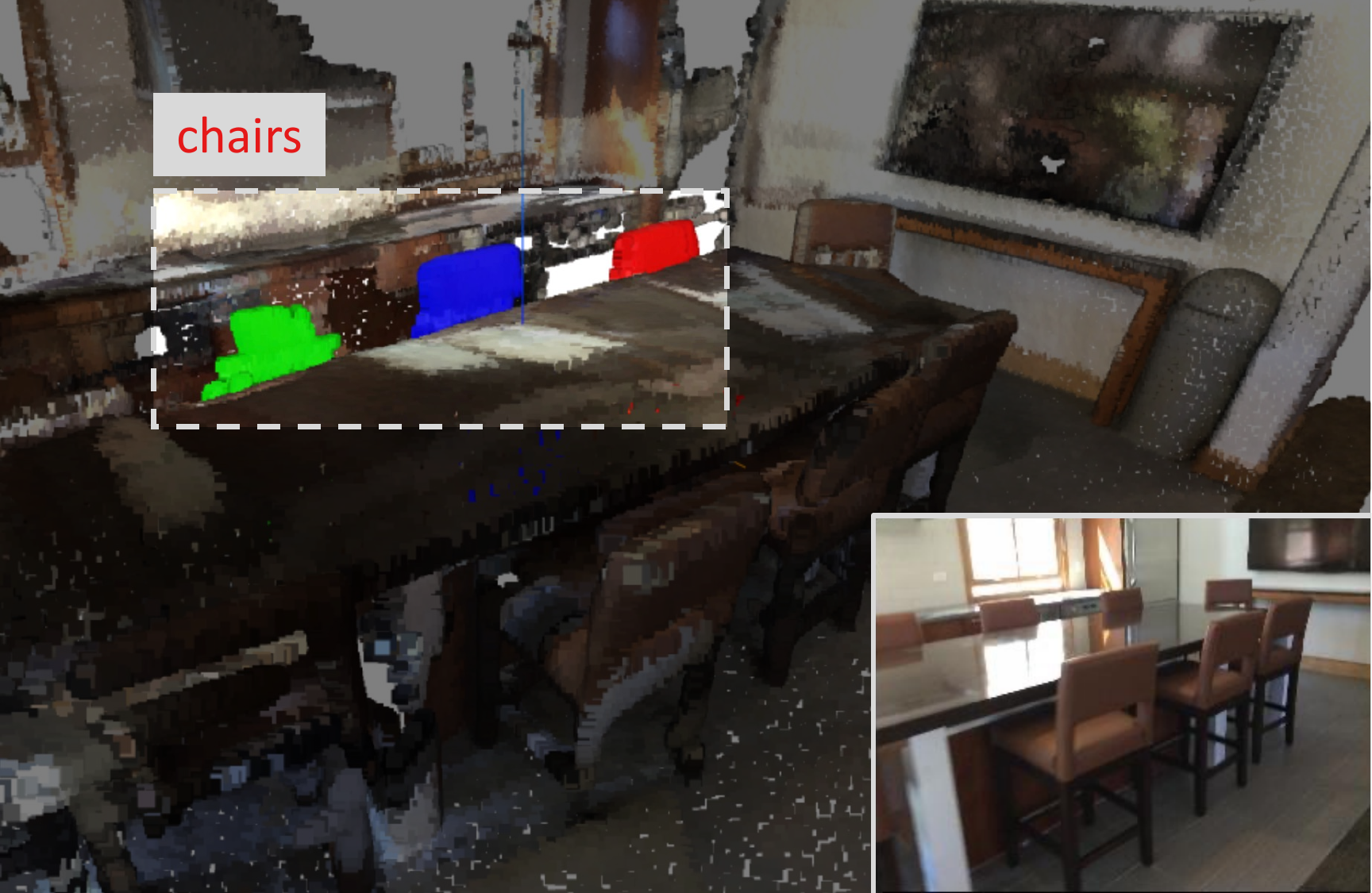}
    \caption{\textit{``Chairs near by the window.''}}
  \end{subfigure}
  \begin{subfigure}{0.323\linewidth}\includegraphics[width=1.0\linewidth]{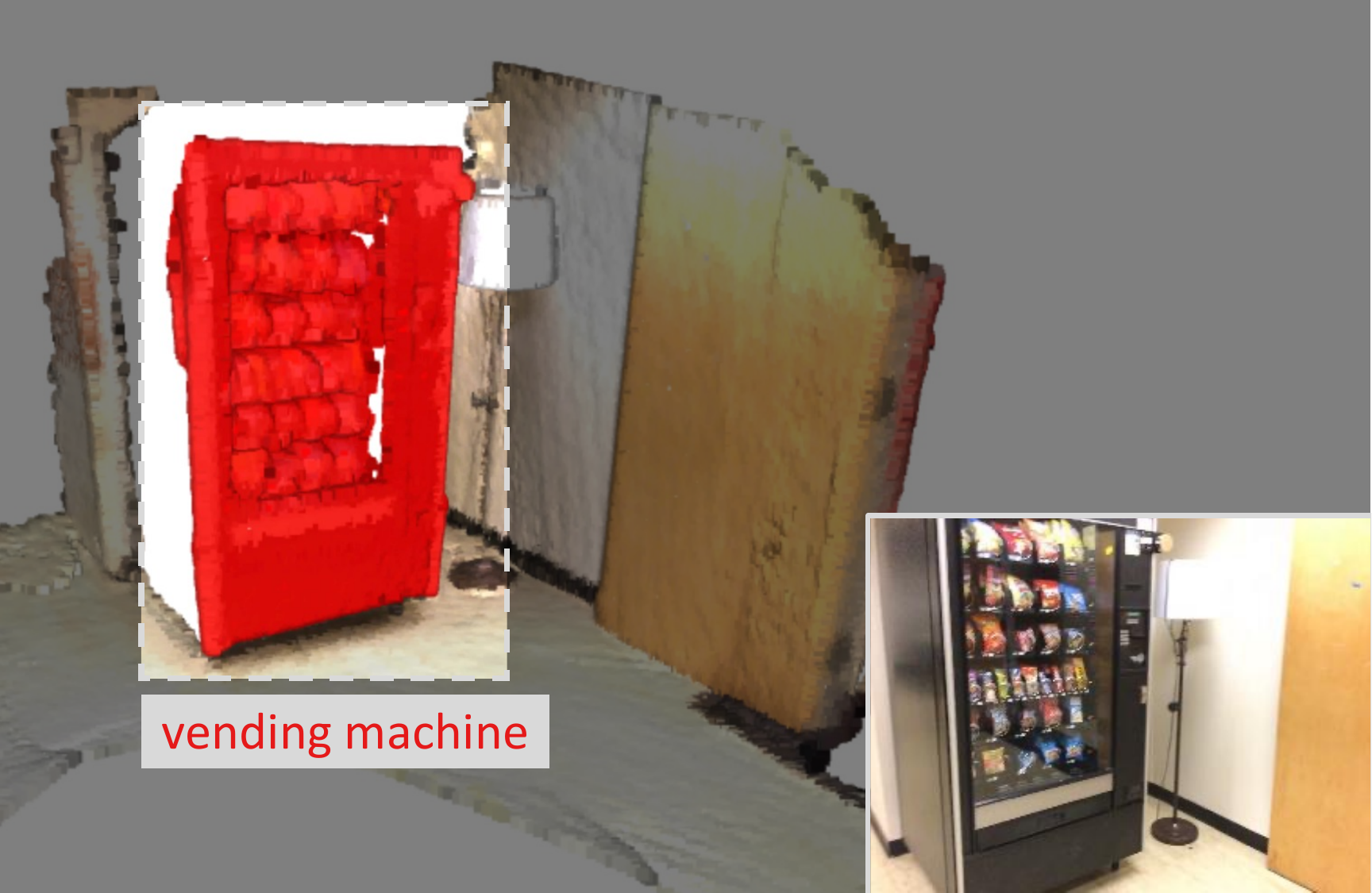}
    \caption{\textit{``I'm hungry.''}}
  \end{subfigure}
  \begin{subfigure}{0.323\linewidth}\includegraphics[width=1.0\linewidth]{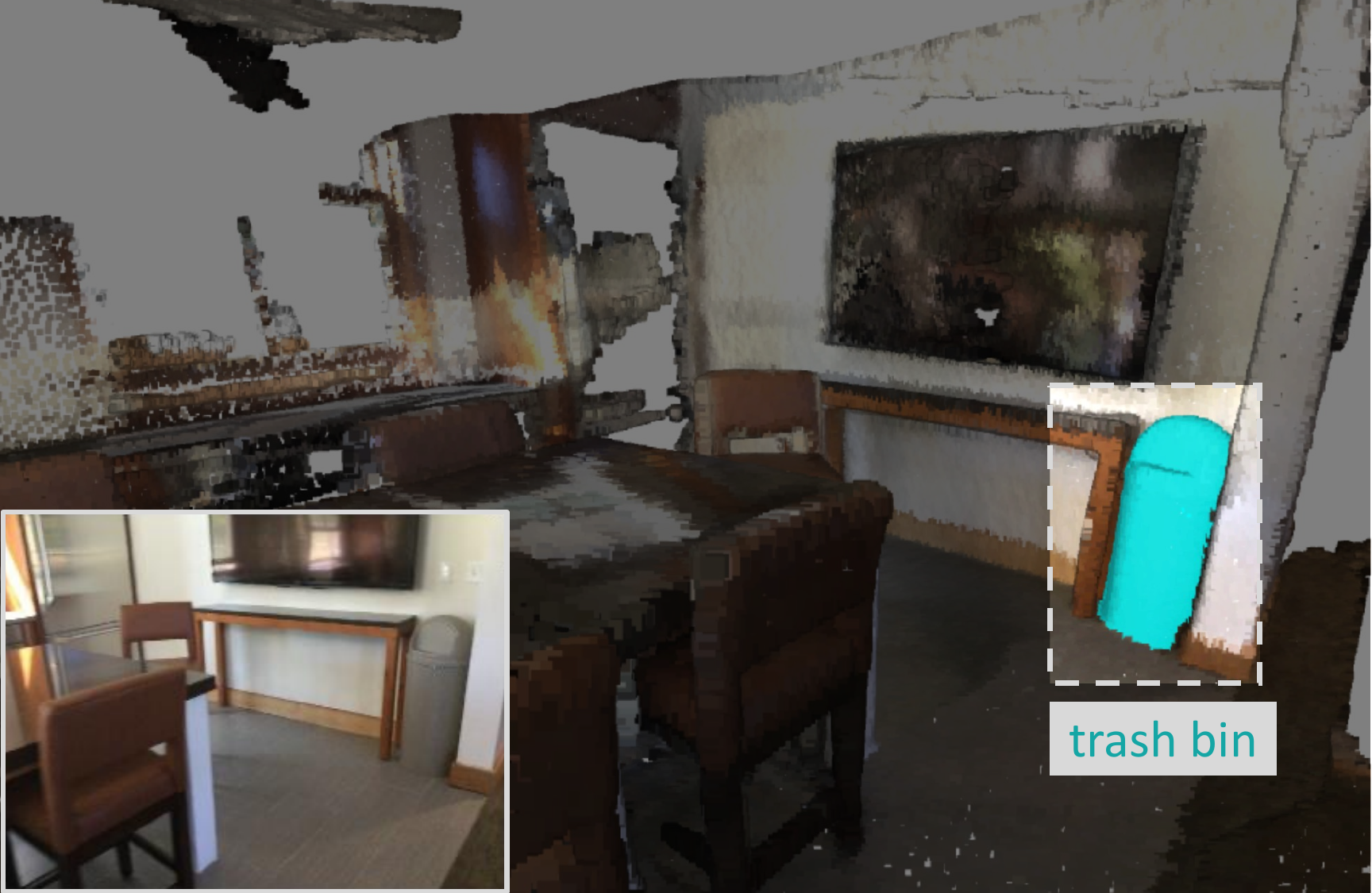}
    \caption{\textit{``Throwing away the garbage.''}}
  \end{subfigure}
  \begin{subfigure}{0.323\linewidth}\includegraphics[width=1.0\linewidth]{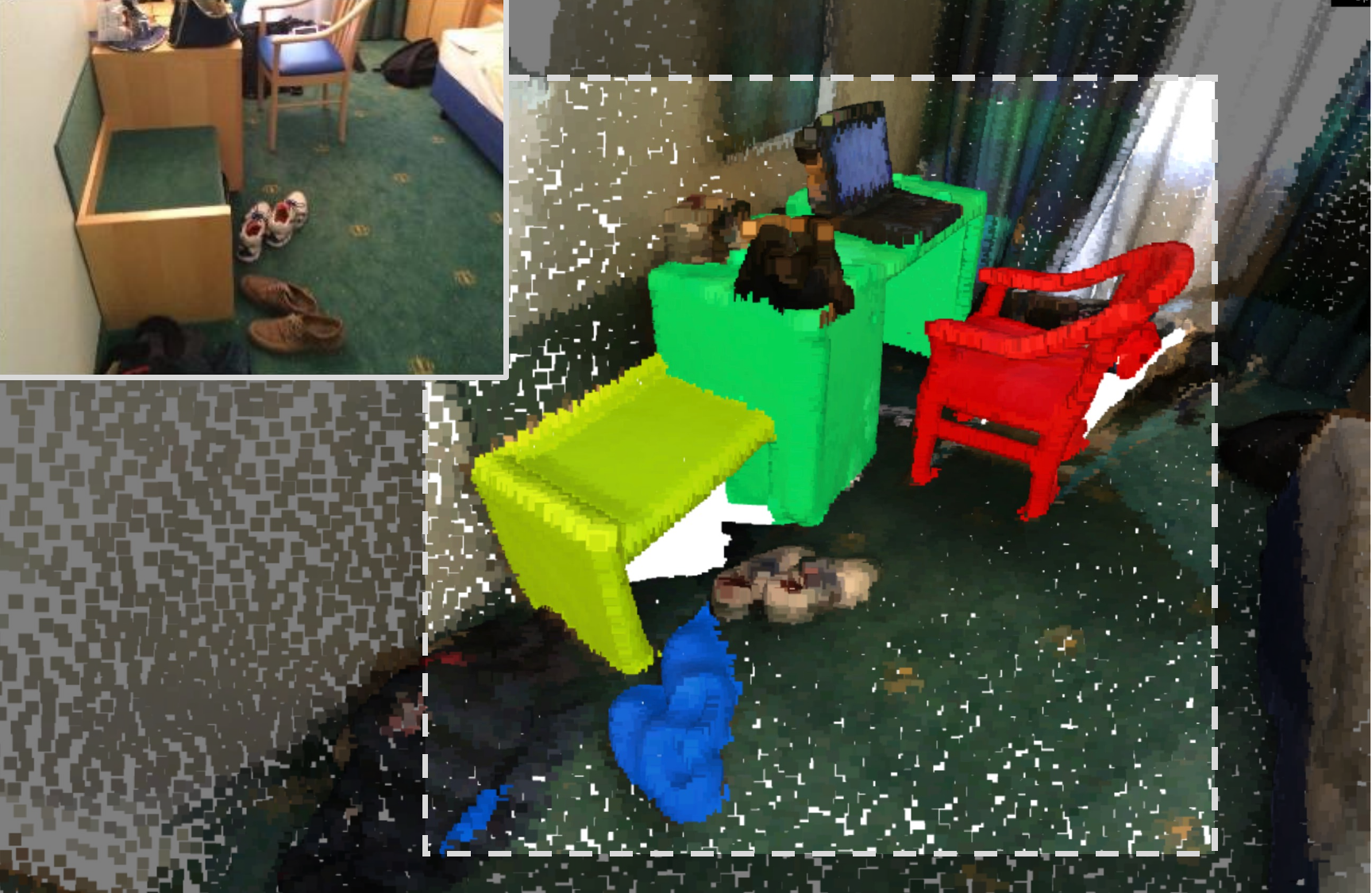}
    \caption{\textit{``Brown furnitures.''}}
  \end{subfigure}
  \begin{subfigure}{0.323\linewidth}\includegraphics[width=1.0\linewidth]{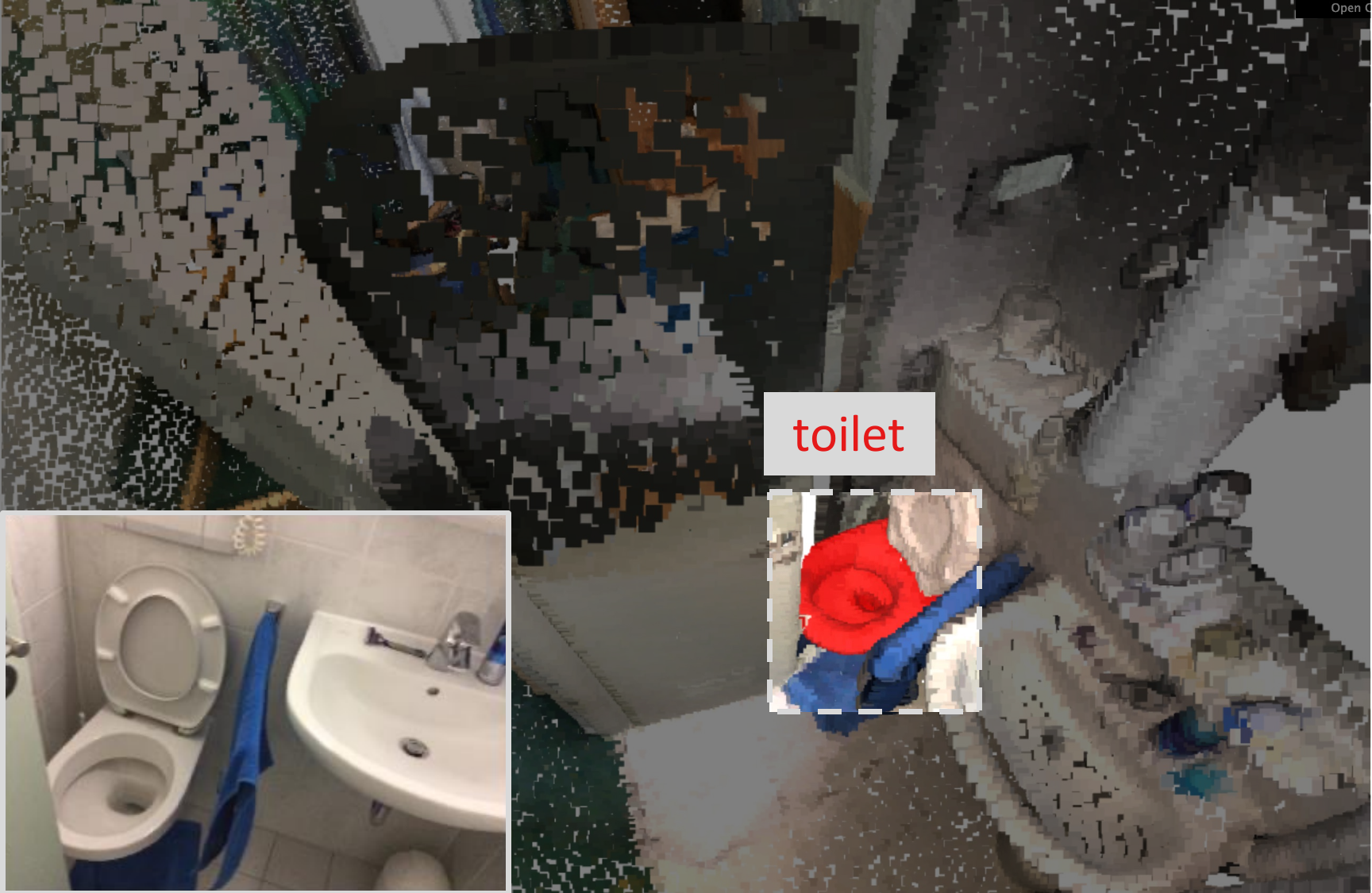}
    \caption{\textit{``Place I can pee.''}}
  \end{subfigure}
  \vspace{-.1in}
  \caption{\textbf{Qualitative results from \ours.} Our \ours demonstrates open-vocabulary capability by effectively responding to free-form language queries, including visual questions, attributes description and functional description.}
  \label{fig:qualitative}
\end{figure}

\begin{figure}[tb]
\captionsetup[subfigure]{}
  \centering
  \begin{subfigure}{0.323\linewidth}\includegraphics[width=1.0\linewidth]{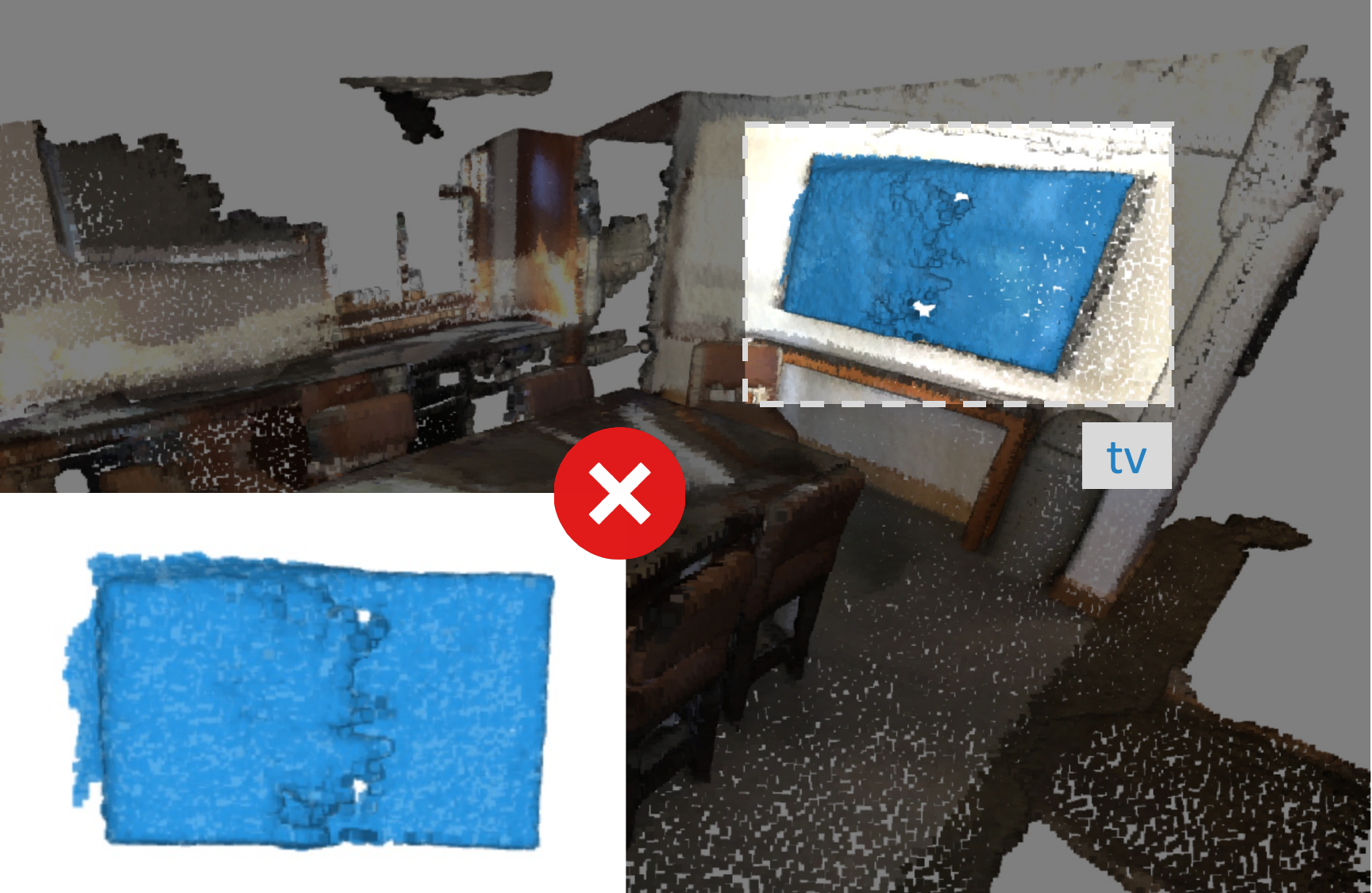}
    \caption{$\mathbf{f}^{MEA}$}
    \label{fig:qa-a}
  \end{subfigure}
  \begin{subfigure}{0.323\linewidth}\includegraphics[width=1.0\linewidth]{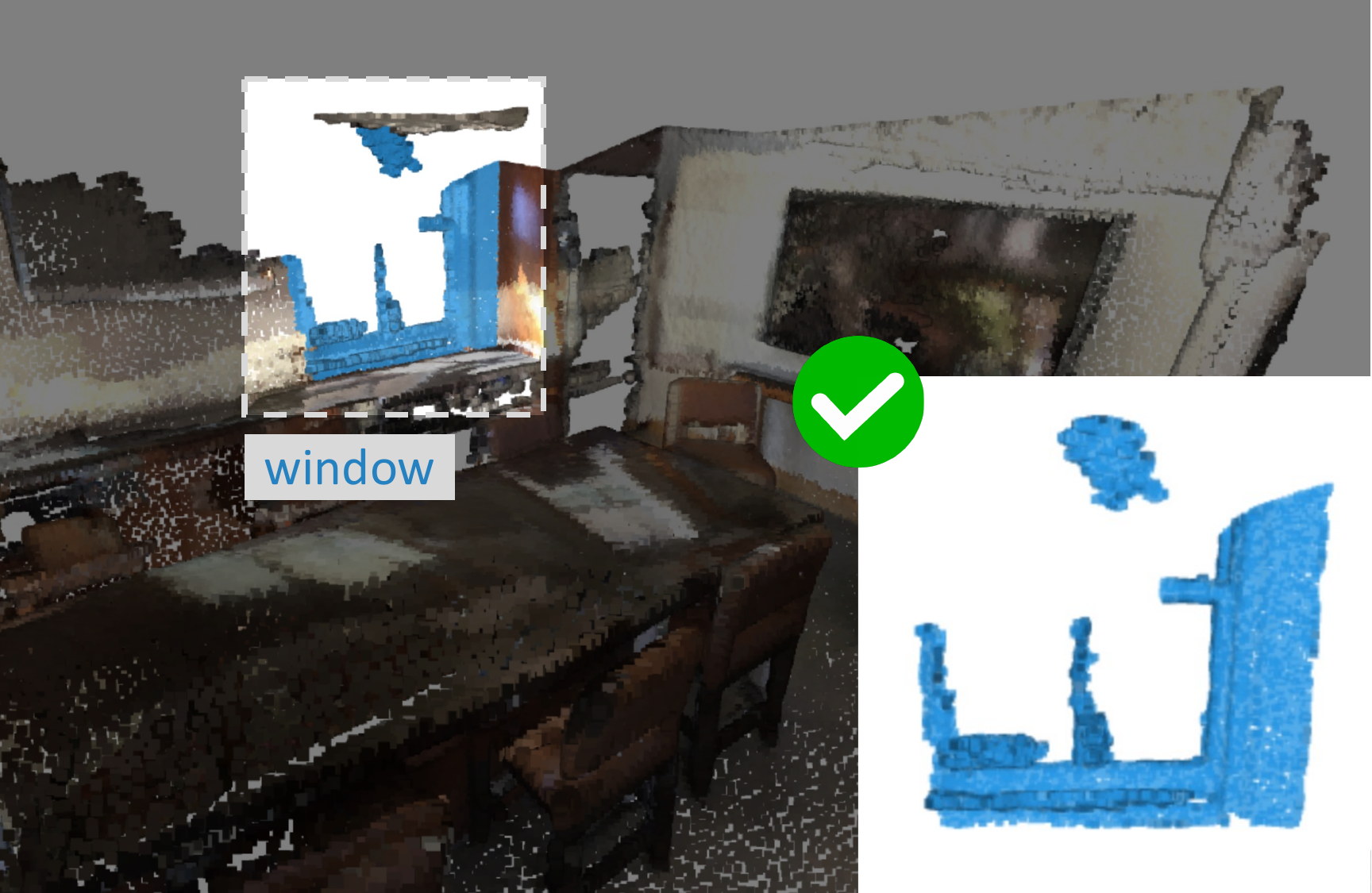}
    \caption{$\mathbf{f}^{MVA}, \mathbf{f}^{MCA}$}
    \label{fig:qa-b}
  \end{subfigure}
  \begin{subfigure}{0.323\linewidth}\includegraphics[width=1.0\linewidth]{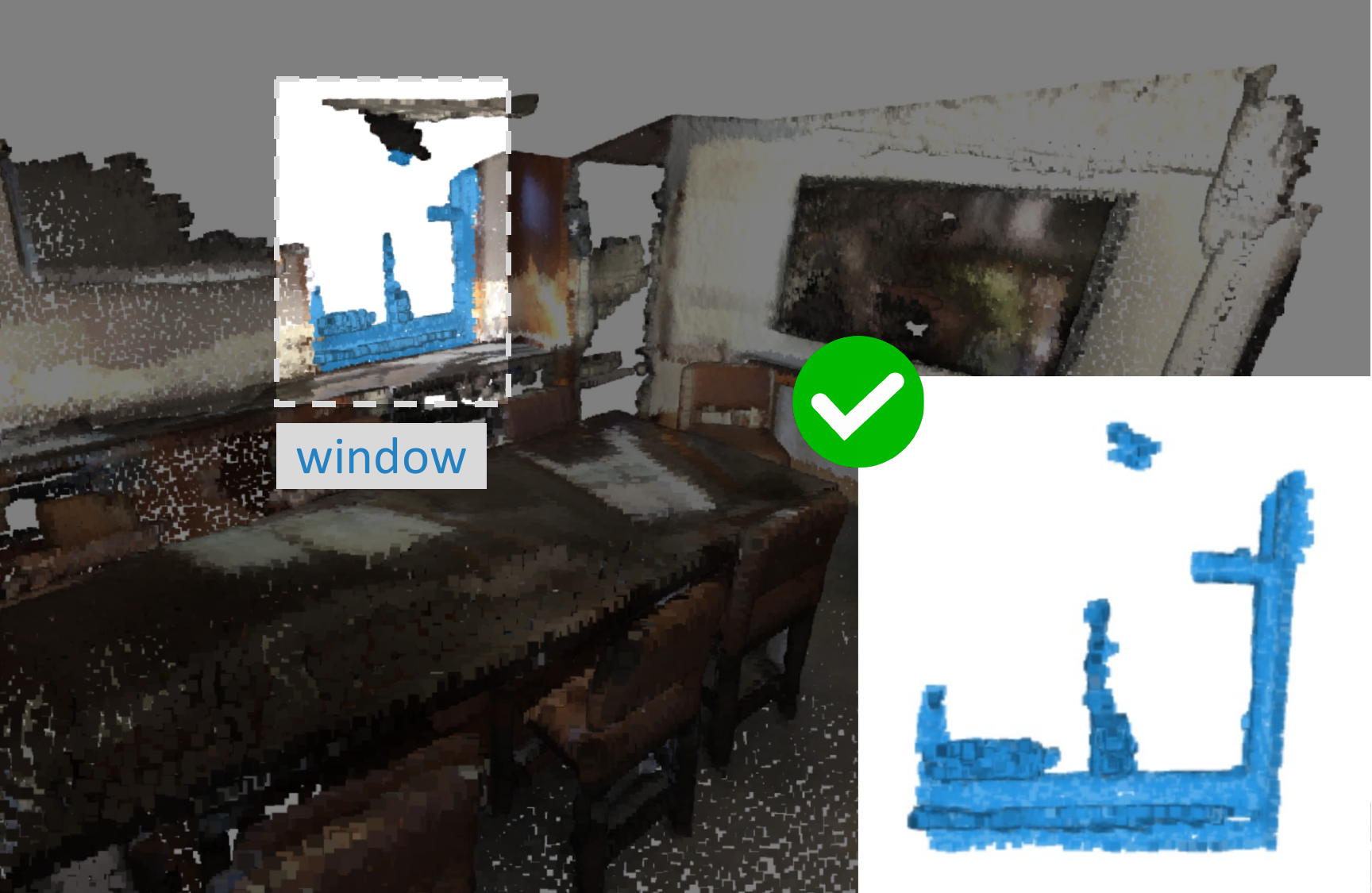}
    \caption{$\mathbf{f}^{MVA}, \mathbf{f}^{MCA},  \mathbf{f}^{MEA}$}
    \label{fig:qa-c}
  \end{subfigure}
  \vspace{-.1in}
  \caption{\textbf{Qualitative analysis on multimodal associations.} Given the free-form language instruction, ``\textit{I wanna see outside.}'', \ours trained only with $\mathbf{f}^{\mathrm{MEA}}$ captures the wrong object ((a)), whereas it segments the related object when $\mathbf{f}^{\mathrm{MVA}}$ and $\mathbf{f}^{\mathrm{MCA}}$ are additionally given as the supervision ((b), (c)).}
  \label{fig:qualitative-association}
  \vspace{-10pt}
\end{figure}

\vspace{2mm}
\noindent \textbf{Multimodal Associations.} 
We analyze the components of multimodal associations ($\mathbf{f}^{\mathrm{MVA}}$, $\mathbf{f}^{\mathrm{MCA}}$, and $\mathbf{f}^{\mathrm{MEA}}$) in Tab.~\ref{tab:multimodal-associations}, reporting the scores of various combinations on ScanNetv2~\cite{dai2017scannet} with 4cm voxel size. We have the following observations. \textbf{First}, using any of multimodal associations can already achieve significant performance, outperforming previous state-of-the-art method (OpenIns3D~\cite{huang2023openins3d}) with larger voxel size (lower resolution). \textbf{Second}, among the three types of associations, mask-entity association $\mathbf{f}^{\mathrm{MEA}}$ is the most effective one on evaluation metrics since it can align the masks with specific categories. \textbf{Third}, when combining $\mathbf{f}^{\mathrm{MEA}}$ with the other two associations, the model suffers from performance degradation on AP and AP$_{50}$ while the performance improves on AP$_{25}$. This observation shows that mask-visual association and mask-caption association can help semantic learning but impair mask accuracy. 
To this end, we further illustrate qualitative results in Fig.~\ref{fig:qualitative-association}.
Given a free-form language instruction instead of category name, \eg, ``I wanna see outside'', the model only using mask-entity association cannot segment the correct instance (Fig.~\ref{fig:qa-a}) while the model incorporating the other associations (Fig.~\ref{fig:qa-b} and Fig.~\ref{fig:qa-c}) can.
Therefore, despite slightly impairing the performance on benchmark, mask-visual association and mask-caption association are crucial to recognizing free-form language instructions, benefiting the applications in real-world scenarios.

\vspace{2mm}
\noindent \textbf{Qualitative Results.} In Fig.~\ref{fig:teaser} and Fig.~\ref{fig:qualitative}, we present qualitative results, demonstrating that \ours is capable of processing free-form language queries, including but not limited to visual questions, attributes description, and functional description.

\section{Conclusion}
In this paper, we propose a novel framework, \ours, for open-vocabulary 3D instance segmentation with free-form language instructions. \ours contains a multimodal fusion network and is supervised with three types of multimodal associations, aiming at aligning the model with various free-form language instructions.
Our framework outperforms previous methods by a large margin on three benchmarks while achieving competitive performance with the fully-supervised counterpart. 
Moreover, extensive qualitative results demonstrate the versatility of our \ours to language instructions.

\clearpage

% \title{Segment Any 3D Object with Language}
% \subtitle{Supplementary Material}
% \maketitle

\appendix
\section*{Appendix}

In this appendix, we provide more implementation details, introduce additional evaluation on free-form language instructions, and conduct more qualitative and quantitative analysis.
\begin{itemize}
    \item More implementation details are provided in Sec.~\ref{sec:details}.
    \item \ours is evaluated on 3D visual grounding task to verify the responsive ability to free-form language instructions in Sec.~\ref{sec:grounding}
    \item Analysis of CLIP visual features are provided in Sec.~\ref{sec:clip-v}.
    \item More qualitative results about the mask caption and segmentation results are shown in Sec.~\ref{sec:quality}.
\end{itemize}

\section{Implementation Details}
\label{sec:details}

\noindent\textbf{Segmentation Network.}
Following Mask3D~\cite{schult2022mask3d}, we use the transformer-based mask-prediction paradigm to obtain instance mask and semantic features. The masks are initialized from object queries and regressed by attention layers. For each 3D point cloud scene, we use farthest point sampling~\cite{qi2017pointnet++} to get 150 points as object queries. After getting masks from the segmentation model, we use DBSCAN~\cite{ester1996density} to break down non-contiguous masks into smaller, spatially contiguous clusters to improve the mask quality. The maximum distance and neighborhood points number are set to $0.95$ and $1$, respectively.

\vspace{2mm}
\noindent\textbf{Text Information Generation and Extraction.}
To effectively generate a caption for each mask, we use a caption model in CLIP space, \ie, DeCap~\cite{li2023decap}. DeCap is a lightweight transformer model to generate captions from CLIP image embedding. It contains a 4-layer Transformer with 4 attention heads as the language model and the visual embedding is obtained from the pre-trained ViT-L/14 CLIP model.
We feed the mask features that are average pooled from the projected CLIP visual features into the DeCap model to obtain the mask caption. Then the caption is integrated into the text prompt ``a \{\} in a scene.'' to better align with our data, \eg ``a blue chair in a scene.''. With the mask caption, noun phrases are extracted by the NLP library, TextBlob~\cite{loria2018textblob} and spaCy~\cite{spacy2}, to get the mask-entity association.

\begin{table}[t]
    \caption{\textbf{Results on ScanRefer~\cite{chen2020scanrefer} for 3D visual grounding task.} \ours achieves the best performance on generalist models with weak supervision.}
    \vspace{-.1in}
    \label{tab:scanrefer}
    \centering
    \begin{tabular}{p{3.5cm}|p{2.5cm}<\centering|p{2cm}<\centering|cc}
    \toprule
    Method  & Type & Supervision & ACC@25 & ACC@50 \\
    \midrule
    OCRand~\cite{chen2020scanrefer} & \multirow{5}{*}{Specialist} & Full & 30.0 & 29.8 \\
    VoteRand~\cite{votenet,chen2020scanrefer} & & Full & 10.0 & 5.3 \\
    SCRC~\cite{scrc} & & Full & 18.7 & 6.5\\
    One-stage~\cite{onestage} & & Full & 20.4 & 9.0\\
    ScanRefer~\cite{chen2020scanrefer}  & & Full & 41.2 & 27.4\\
    \midrule
    3D-LLM (flamingo)~\cite{3dllm} & \multirow{2}{*}{Generalist} & Full & 21.2 & --- \\
    \textbf{\ours} & & \textbf{Weak} & \textbf{25.2} & \textbf{22.6}\\
    \bottomrule
    \end{tabular}
\end{table}

\section{3D Visual Grounding}
\label{sec:grounding}
To further verify the effectiveness of \ours on various language instructions, we conduct experiments on 3D visual grounding benchmark ScanRefer~\cite{chen2020scanrefer}. 3D visual grounding aims at localizing 3D objects with free-form text descriptions. Therefore, we query \ours with each text prompt in the ScanRefer validation set to get the corresponding instance, and then 3D bounding box is obtained from the instance masks. The performance is evaluated on the matching accuracy with IoU over 0.25 (ACC@25) and 0.5 (ACC@50).

\noindent\textbf{Baselines.}
We compare \ours with five specialist baseline models and one generalist model. Specialist models mean that the models are designed and trained for 3D visual grounding only while generalist models are models that can address other tasks such as instance segmentation with class names.
For the specialist models,
OCRand~\cite{chen2020scanrefer} uses an oracle with ground truth bounding boxes of objects, and selects a random box that matches the object category. 
VoteRand~\cite{votenet,chen2020scanrefer} leverages the pre-trained VoteNet~\cite{votenet} to predict bounding boxes and randomly select a box of the correct semantic class.
SCRC~\cite{scrc} and One-stage~\cite{onestage} are 2D approaches with 3D extension using back-projection. ScanRefer~\cite{chen2020scanrefer} uses a pre-trained VoteNet~\cite{votenet} with a trained GRU to select a matching bounding box.
For the generalist model, 3D-LLM directly predicts the location of the bounding box corresponding to the text description via large language model.
\textbf{Note that} except for OCRand and VoteRand %that
where training is not required, the other four baseline models are trained or fine-tuned on the ScanRefer training set. Differently, \ours \textbf{only use mask annotation} of ScanNetv2~\cite{dai2017scannet} training set and the text description in ScanRefer are totally ignored during training.

\begin{sloppypar}
\noindent\textbf{Results.}
As shown in Tab.~\ref{tab:scanrefer}, with only mask annotation, \ours outperforms the generalist model 3D-LLM by 4\% on ACC@25. In addition, \ours can achieve competitive performance with the fully-supervised specialist counterpart on ACC@50 (22.6\% \textit{v.s.} 27.4\%). 
Such results demonstrate the strong generalization ability and the effectiveness in responding to free-form language instructions of our framework.
\end{sloppypar}

\section{Analysis of CLIP Visual Feature}
\label{sec:clip-v}
CLIP visual features play an important role in the generalization ability of \ours. In this section, we further analyze the effectiveness of CLIP visual features in the backbone feature ensemble and inference ensemble.

\subsection{Backbone Feature Ensemble}

\begin{figure}[t]
\captionsetup[subfigure]{}
  \centering
  \begin{subfigure}{0.3233\linewidth}\includegraphics[width=1.0\linewidth]{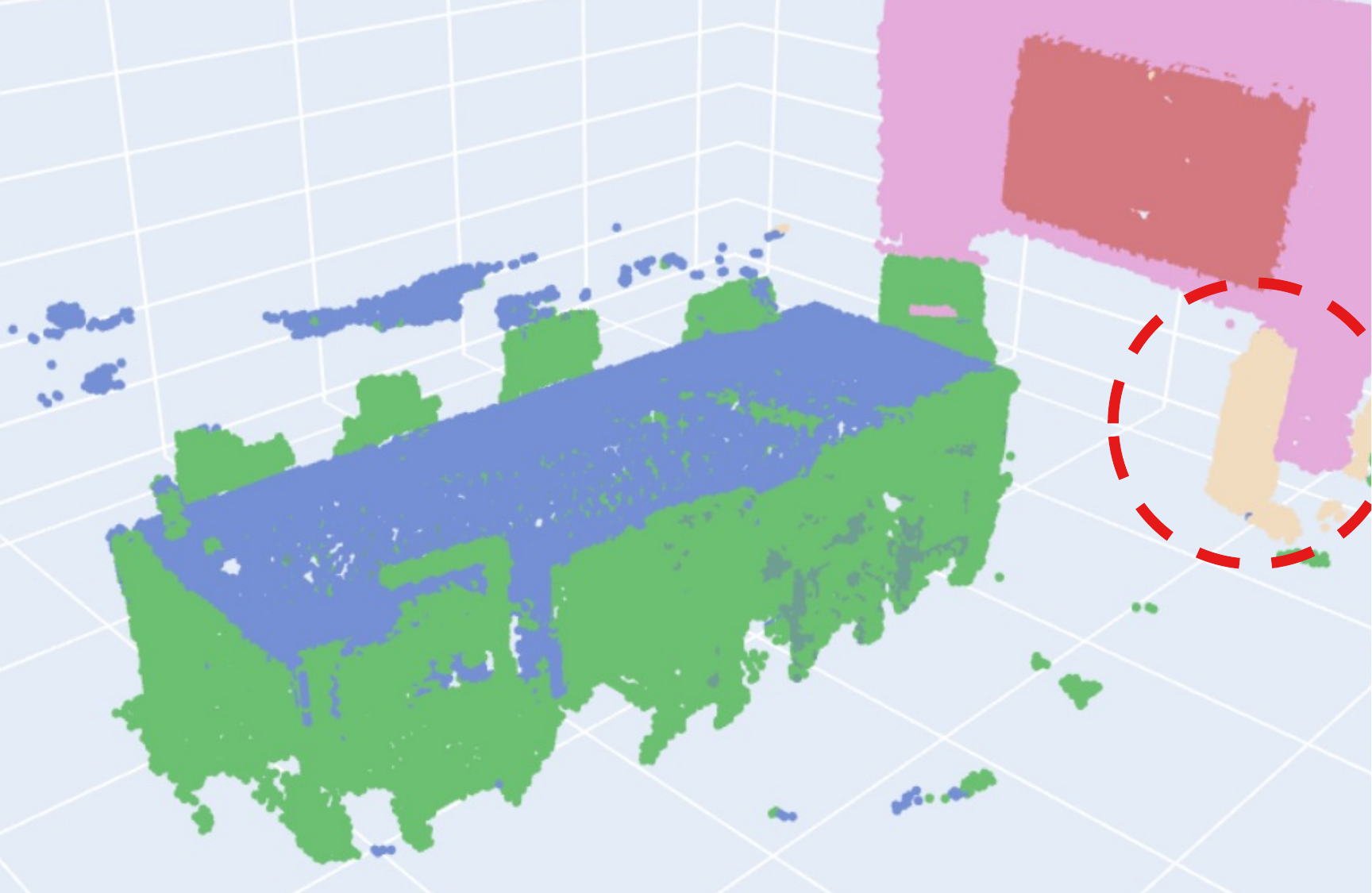}
    \caption{K-Means clustering of $\mathbf{f}^{p}$}
    \label{fig:kmean-a}
  \end{subfigure}
  \begin{subfigure}{0.3233\linewidth}\includegraphics[width=1.0\linewidth]{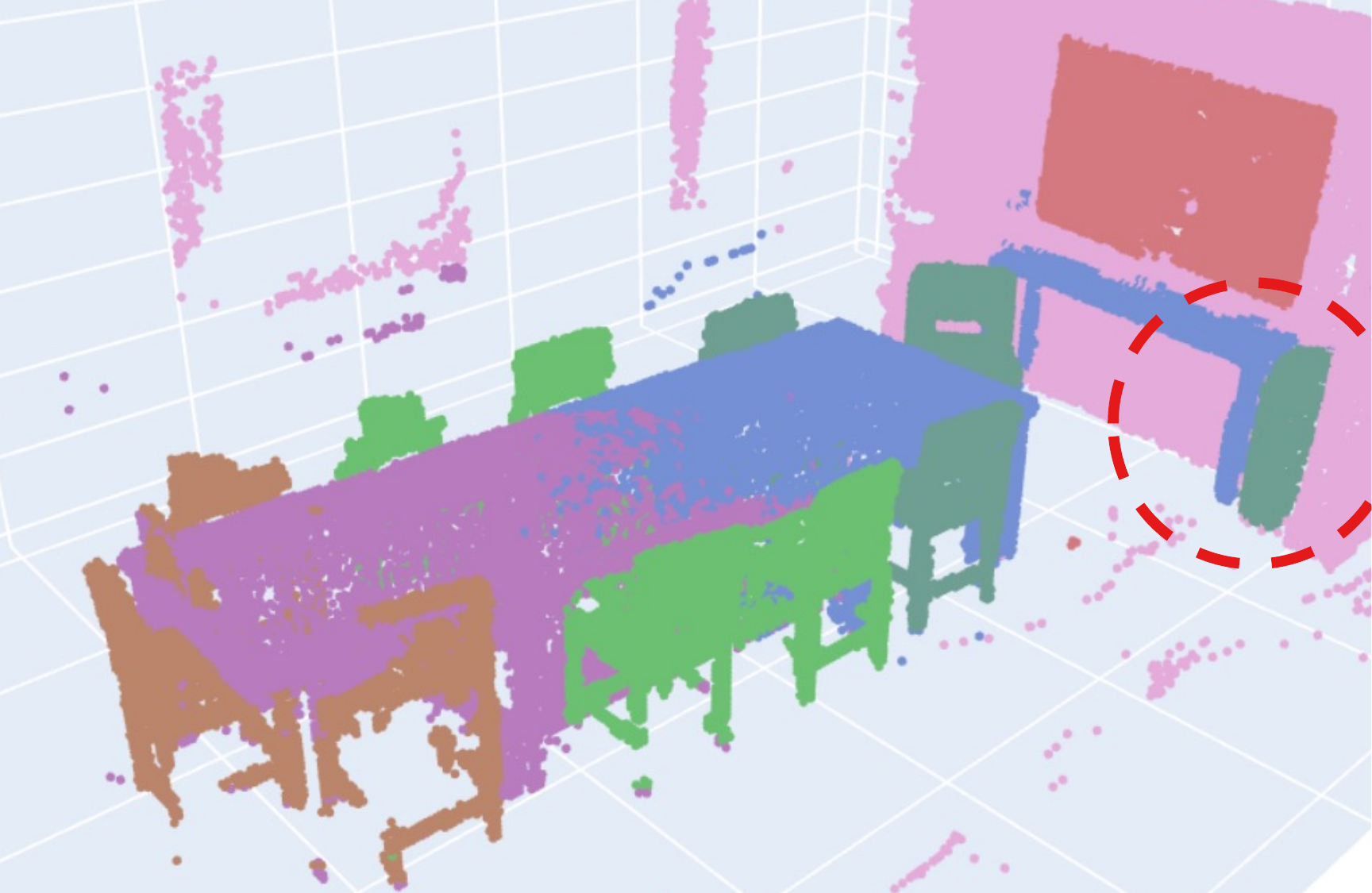}
    \caption{K-Means clustering of $\mathbf{f}^{b}$}
    \label{fig:kmean-b}
  \end{subfigure}
  \begin{subfigure}{0.3233\linewidth}\includegraphics[width=1.0\linewidth]{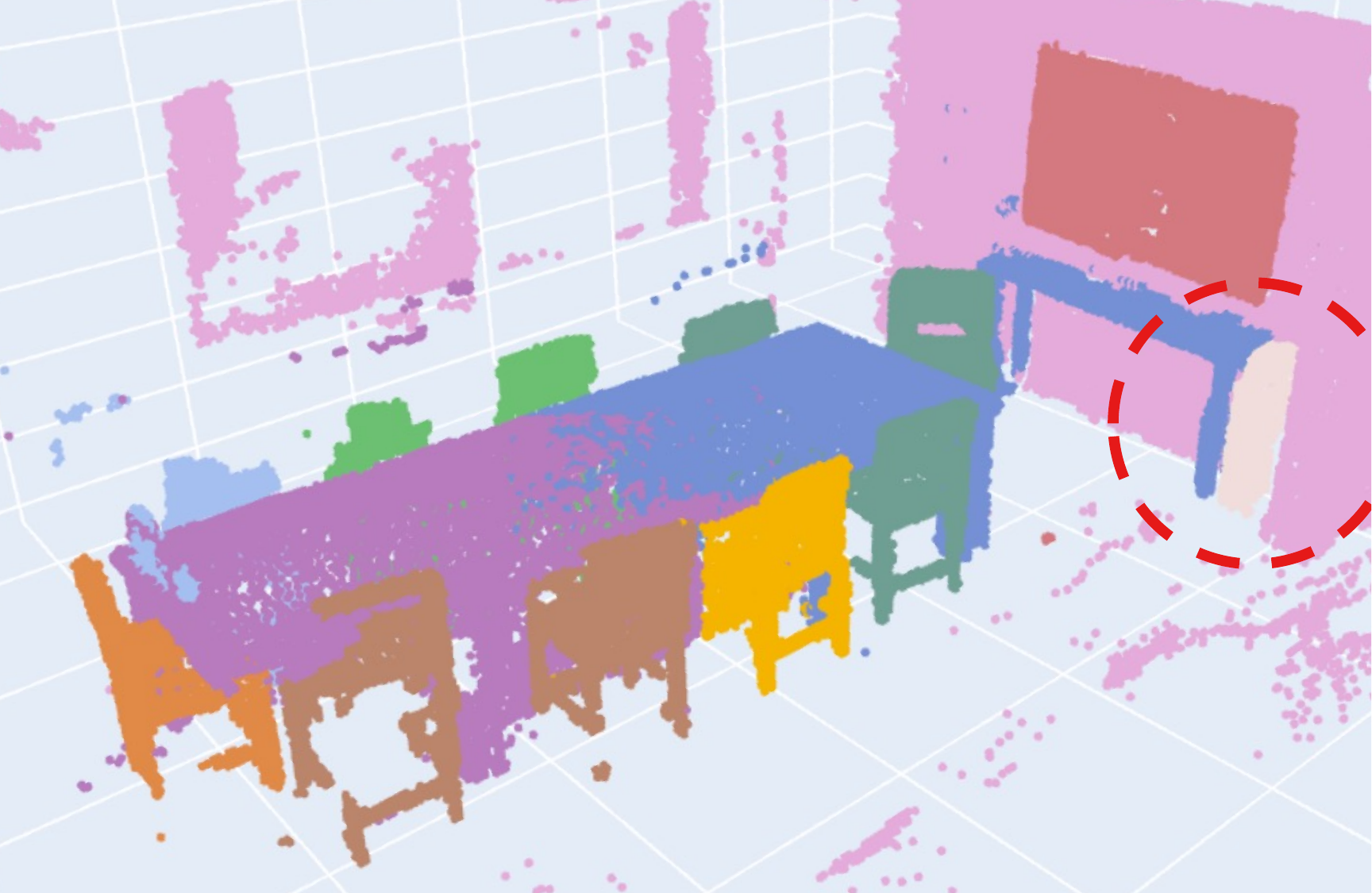}
    \caption{K-Means clustering of $\tilde{\mathbf{f}}^{b}$}
    \label{fig:kmean-c}
  \end{subfigure}
  \vspace{-.1in}
  \caption{\textbf{K-means clustering of different backbone features.} 
  Different colors denote different clusters.}
  \label{fig:kmean}
\end{figure}

\begin{table}[t]
  \caption{\textbf{Analysis on classification probability ensemble.} 
  % Different methods to combine the probabilities from 3D segmentation model and CLIP model are investigated. 
  Results are reported on the ScanNetv2~\cite{dai2017scannet} dataset in 2cm voxel size.
  }
  \vspace{-.1in}
  \label{tab:logit}
  \centering
  \small
  \begin{tabularx}{0.8\textwidth}
  { 
   >{\hsize=1.4\hsize}X |
  >{\centering\arraybackslash\hsize=.32\hsize\linewidth=\hsize}X 
  >{\centering\arraybackslash\hsize=.32\hsize\linewidth=\hsize}X 
  >{\centering\arraybackslash\hsize=.32\hsize\linewidth=\hsize}X |
  >{\centering\arraybackslash\hsize=.5\hsize\linewidth=\hsize}X}
    \toprule
     Component & AP & AP$_{50}$ & AP$_{25}$ & voxel size \\
    \midrule
   w.o. Ensemble & 42.2 & 58.6 & 66.9 & 2cm \\
   hard geometric mean & 43.7 & 61.1 & 70.1 & 2cm \\
   soft geometric mean (\textit{ours}) & \textbf{44.4} & \textbf{62.2} & \textbf{71.4} & 2cm \\
  \bottomrule
  \end{tabularx}
\end{table}

\begin{figure}[t]
\captionsetup[subfigure]{}
  \centering
  \begin{subfigure}{0.3233\linewidth}\includegraphics[width=1.0\linewidth]{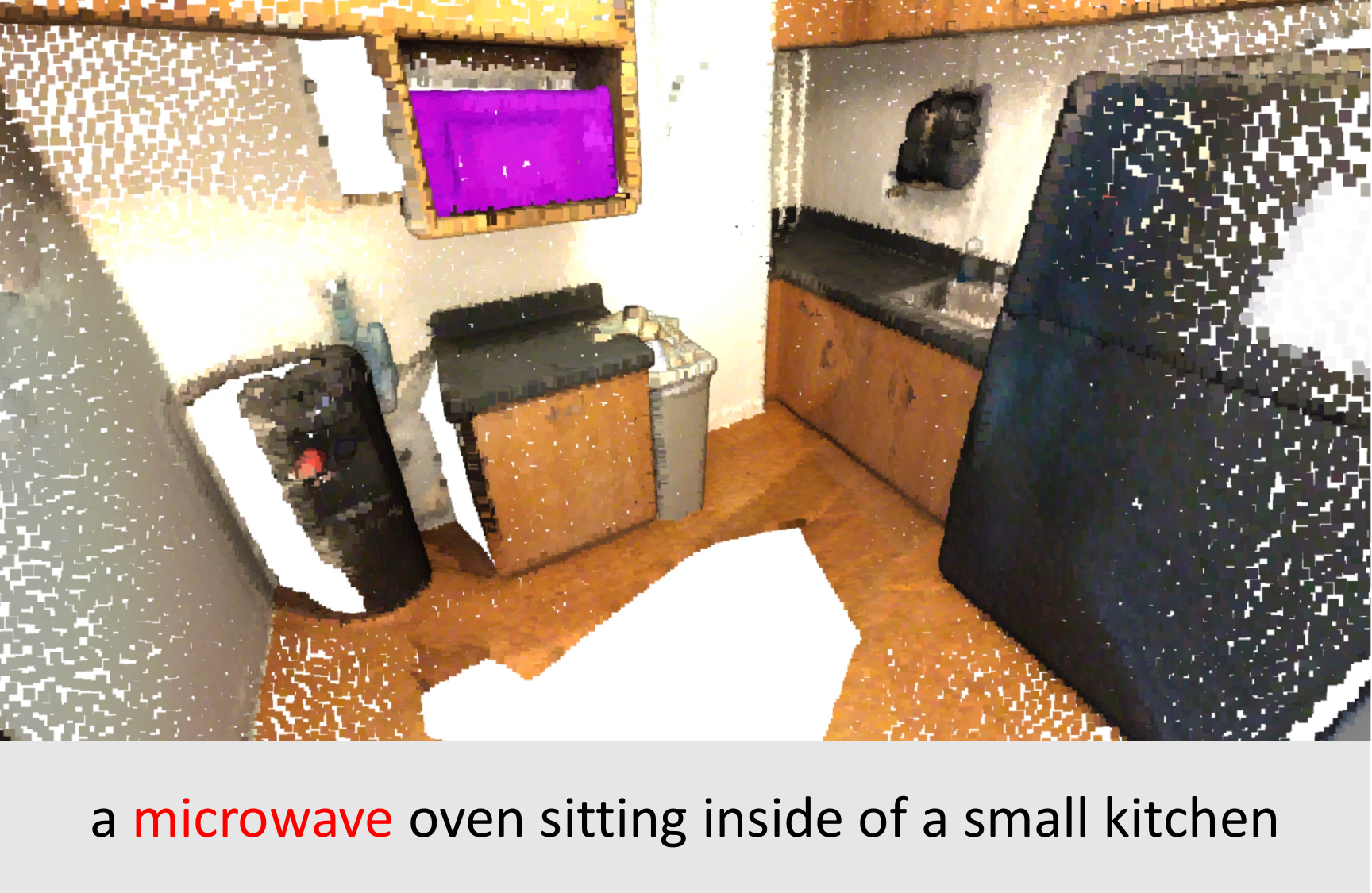}
  \end{subfigure}
  \begin{subfigure}{0.3233\linewidth}\includegraphics[width=1.0\linewidth]{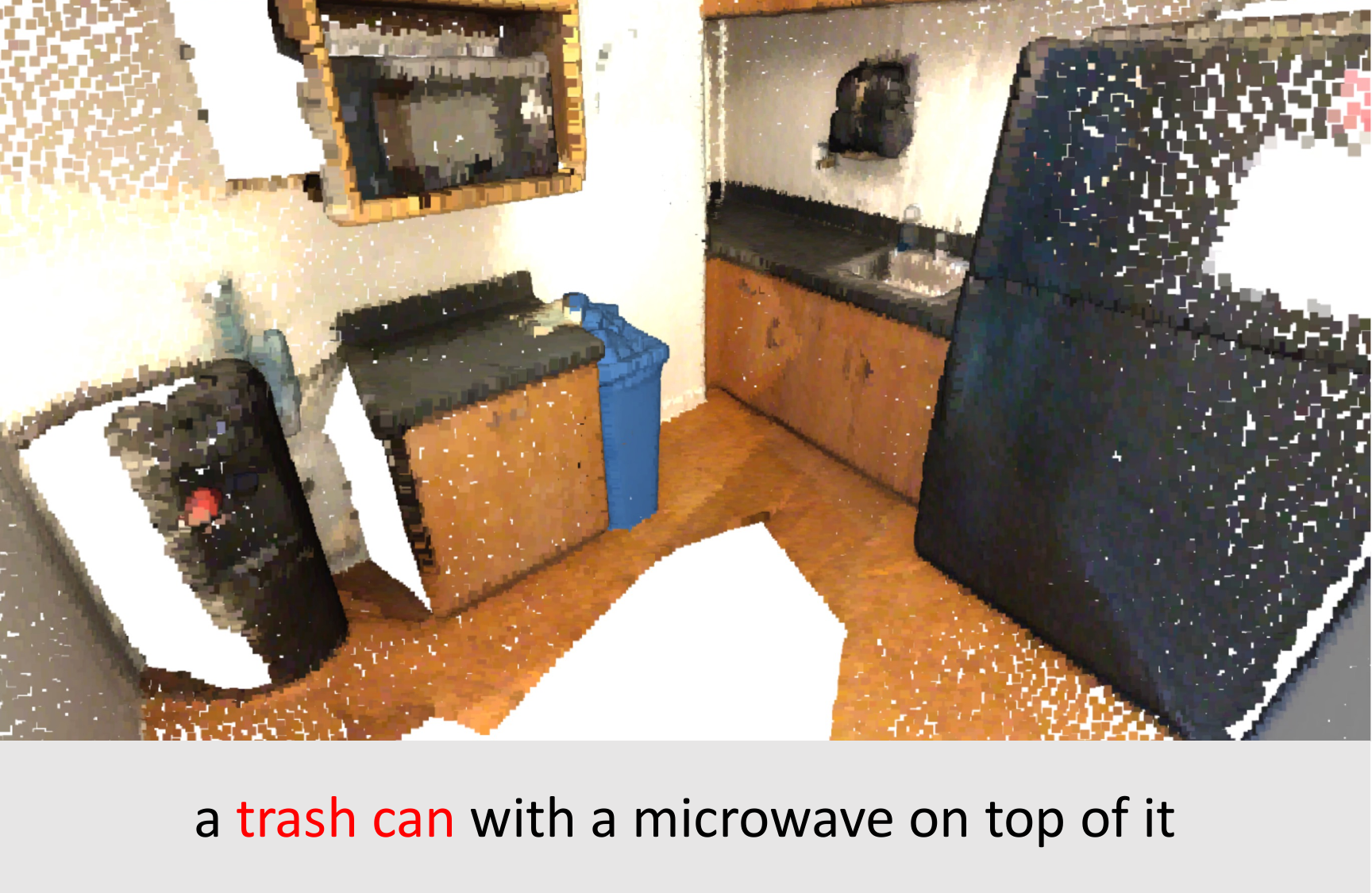}
  \end{subfigure}
  \begin{subfigure}{0.3233\linewidth}\includegraphics[width=1.0\linewidth]{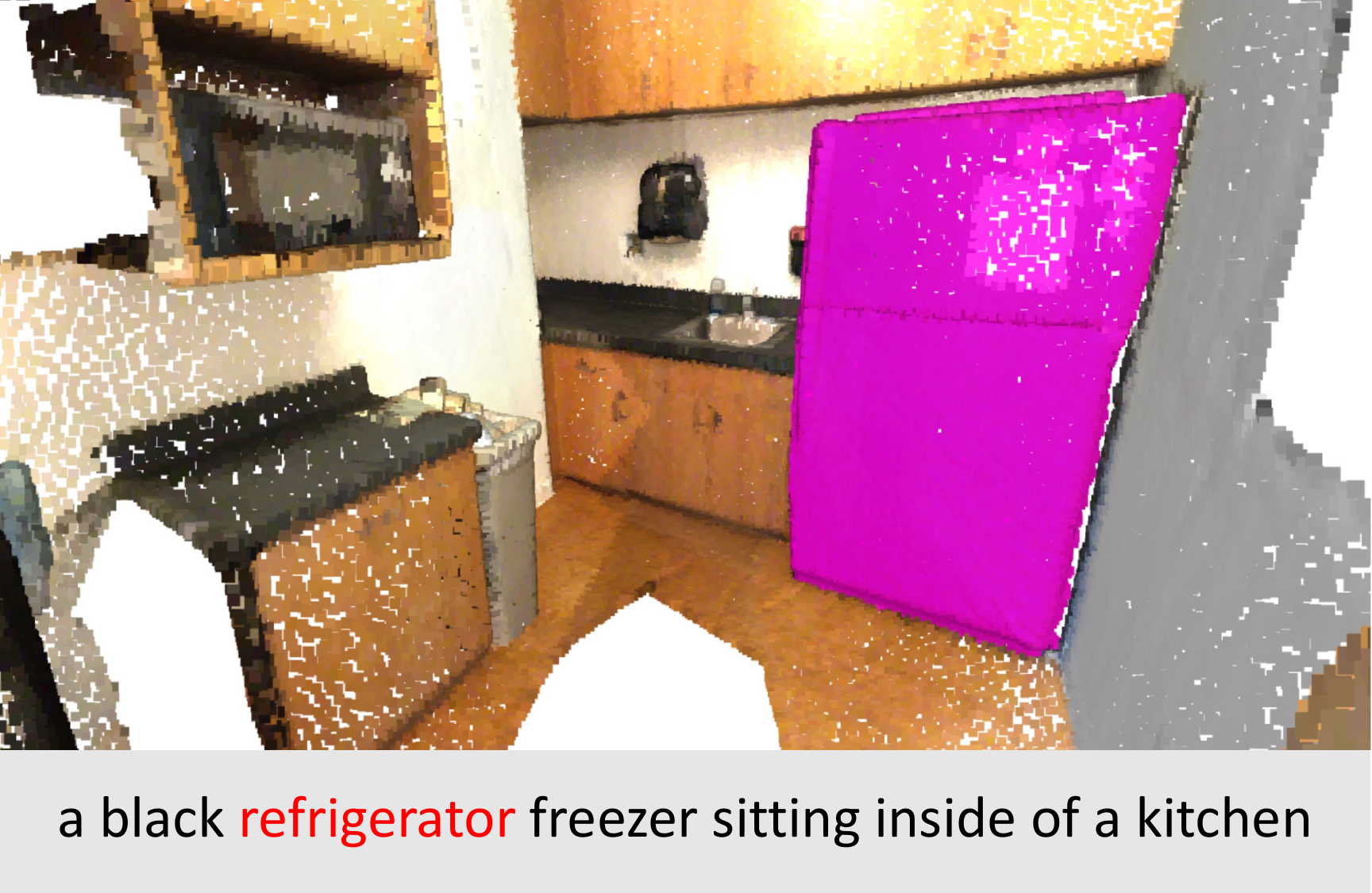}
  \end{subfigure}
  \begin{subfigure}{0.3233\linewidth}\includegraphics[width=1.0\linewidth]{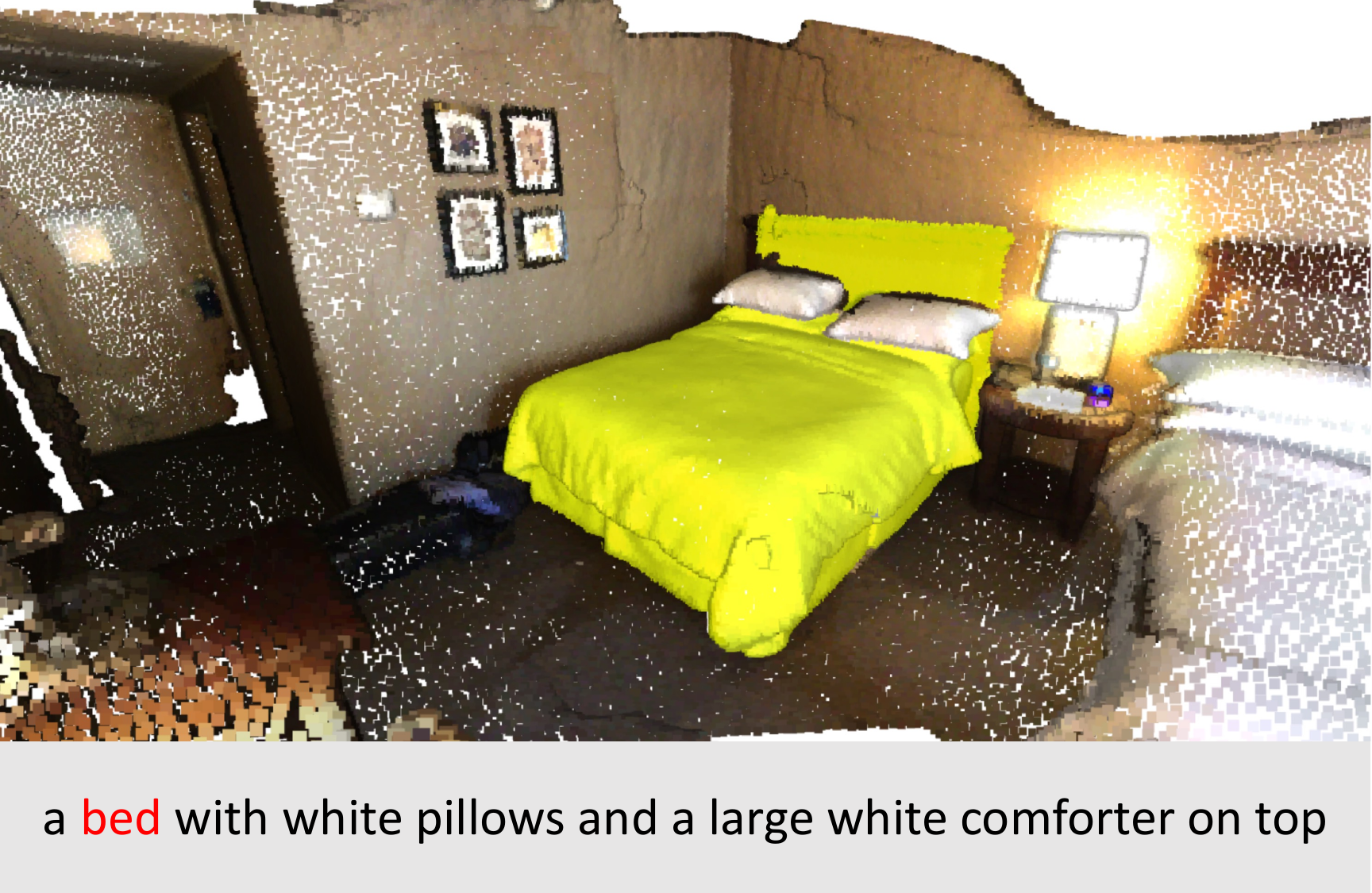}
  \end{subfigure}
  \begin{subfigure}{0.3233\linewidth}\includegraphics[width=1.0\linewidth]{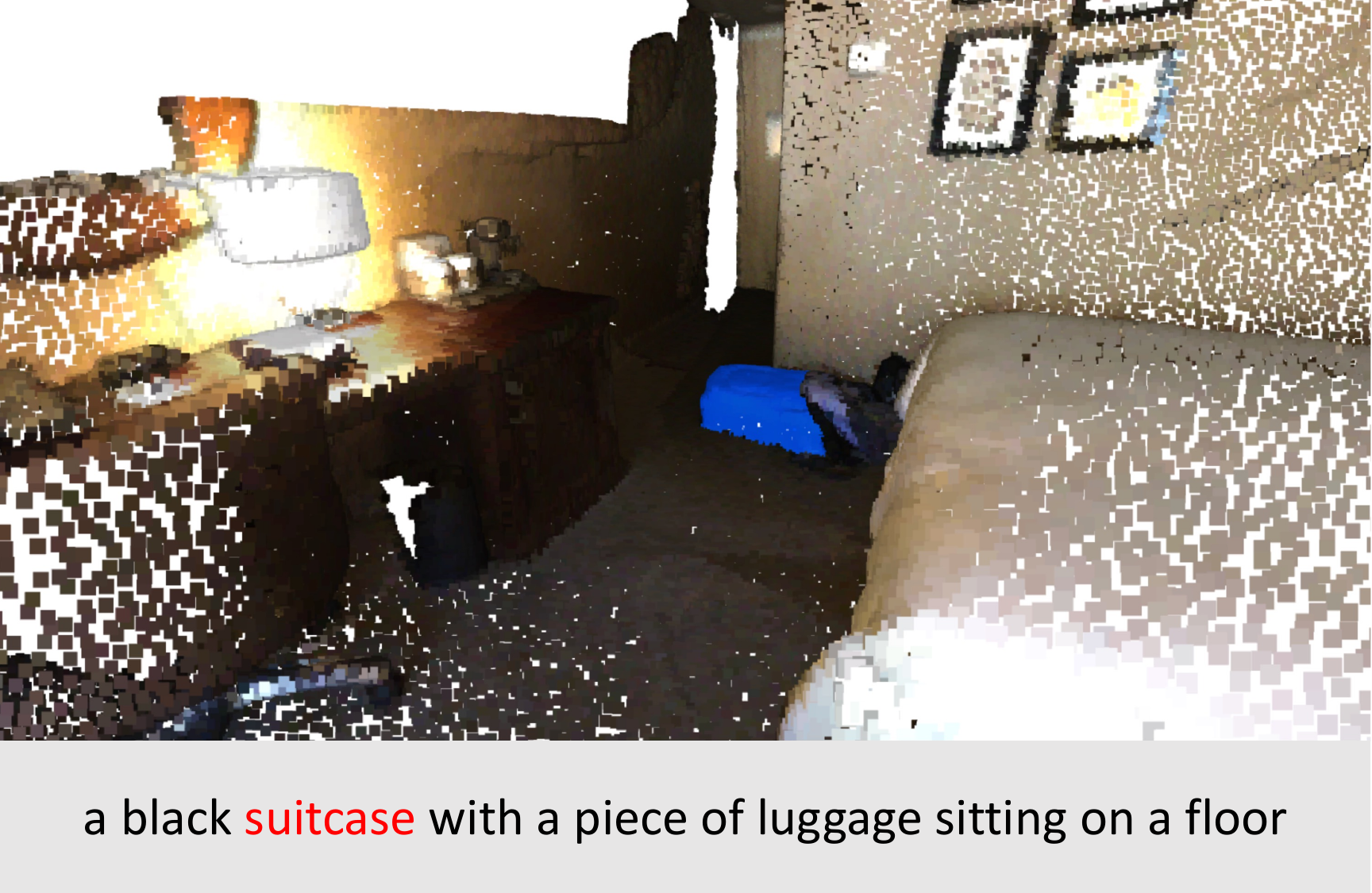}
  \end{subfigure}
  \begin{subfigure}{0.3233\linewidth}\includegraphics[width=1.0\linewidth]{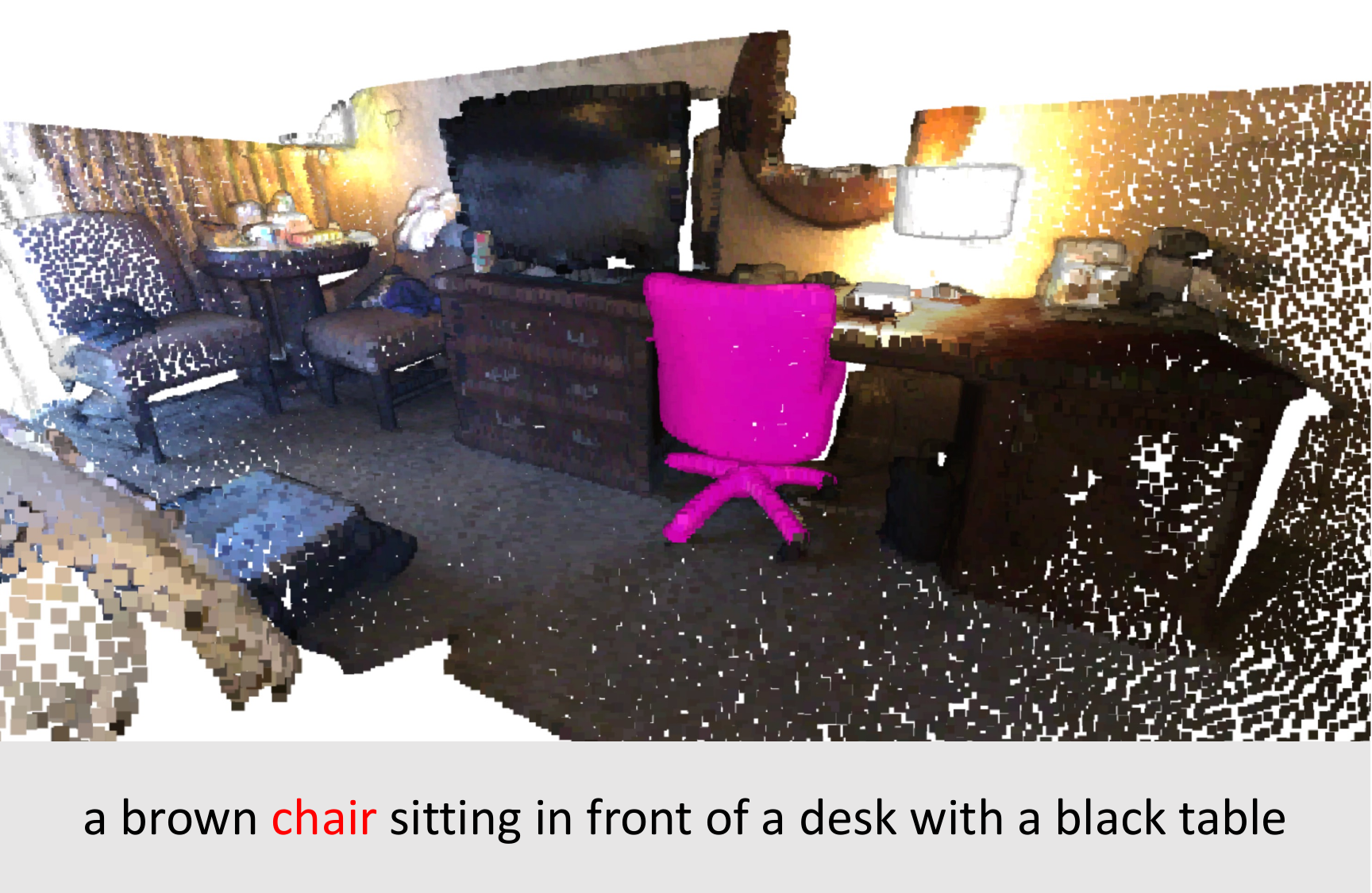}
  \end{subfigure}
  \begin{subfigure}{0.3233\linewidth}\includegraphics[width=1.0\linewidth]{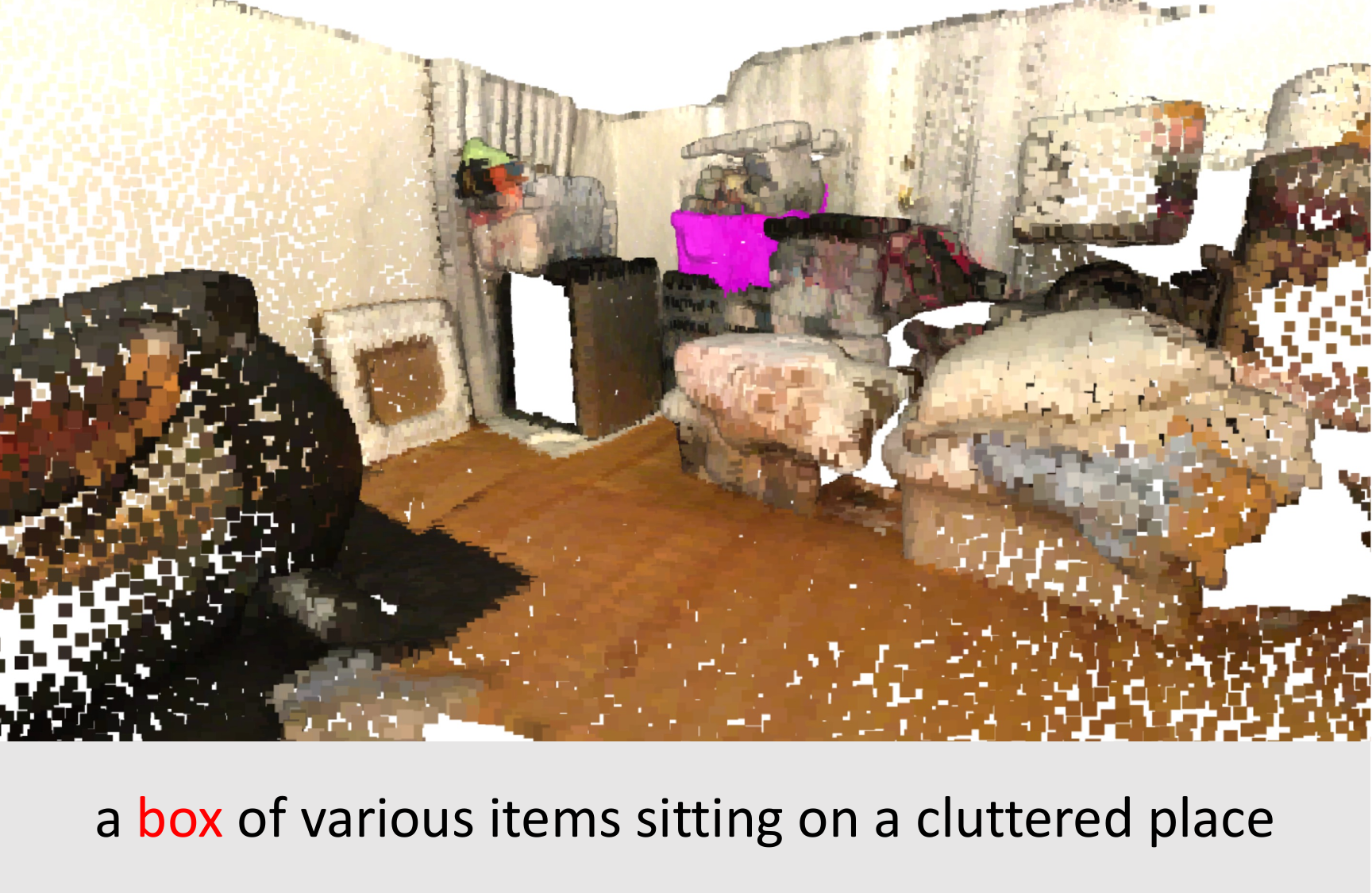}
  \end{subfigure}
  \begin{subfigure}{0.3233\linewidth}\includegraphics[width=1.0\linewidth]{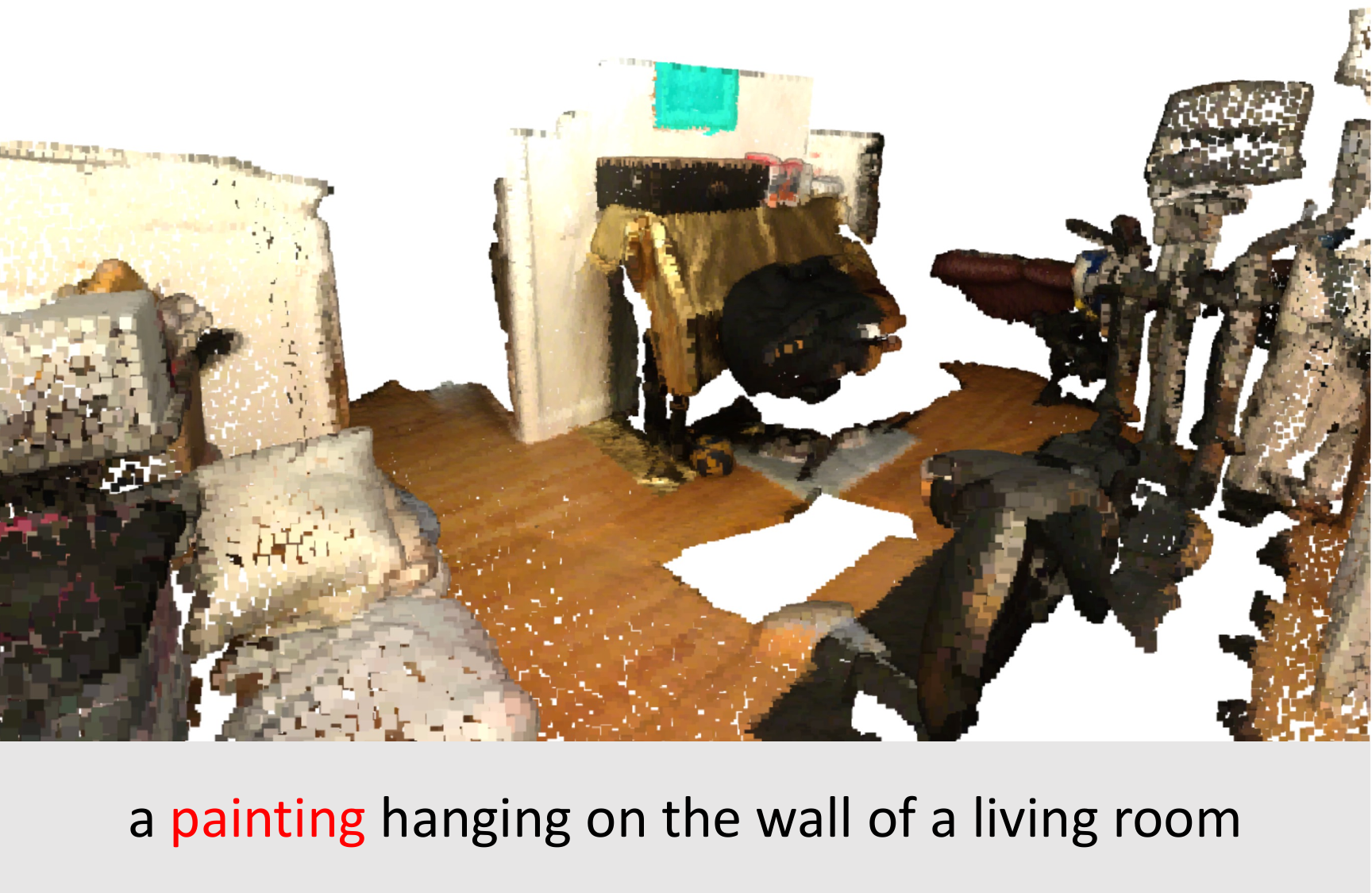}
  \end{subfigure}
  \begin{subfigure}{0.3233\linewidth}\includegraphics[width=1.0\linewidth]{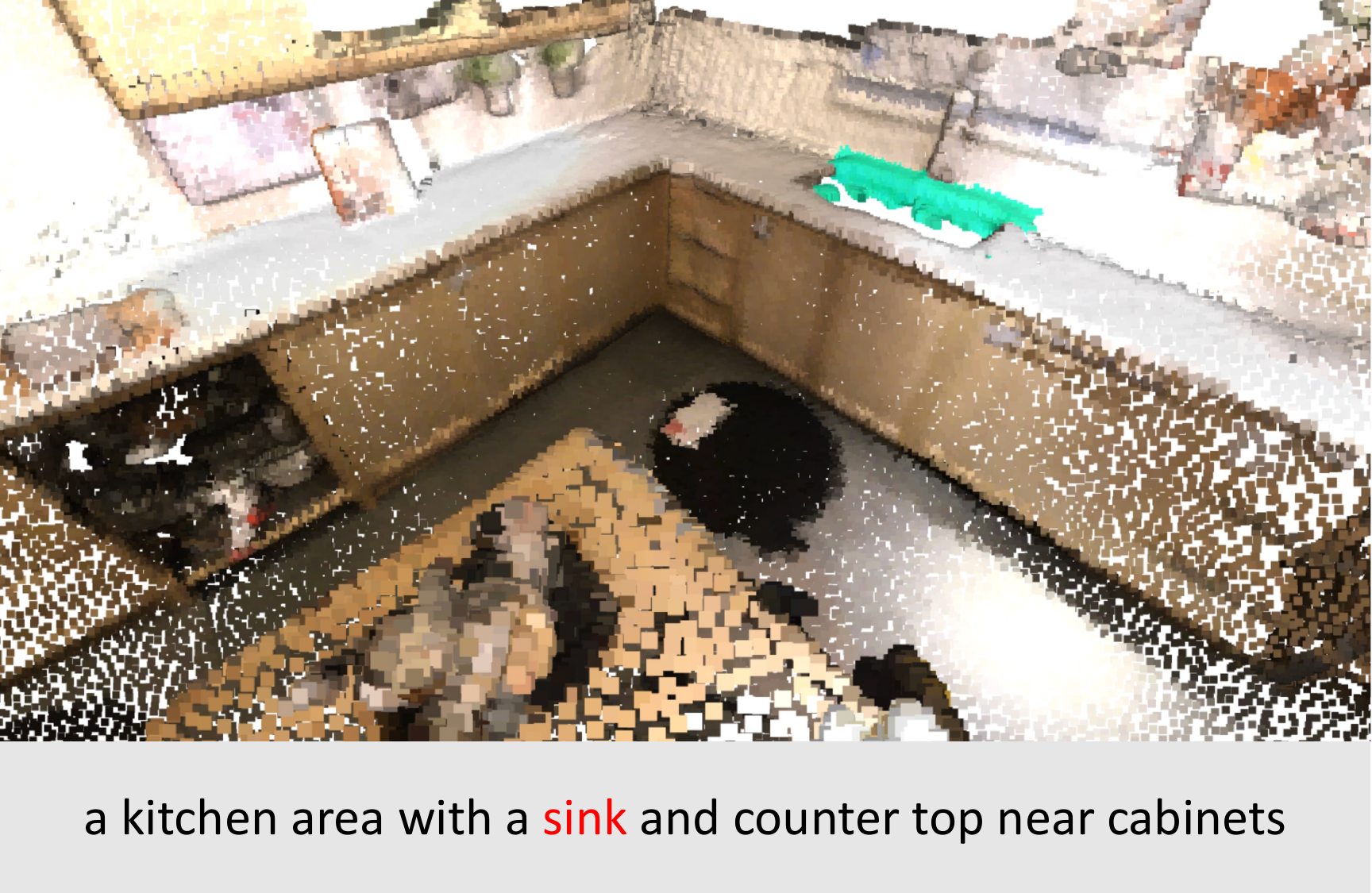}
  \end{subfigure}
  \begin{subfigure}{0.3233\linewidth}\includegraphics[width=1.0\linewidth]{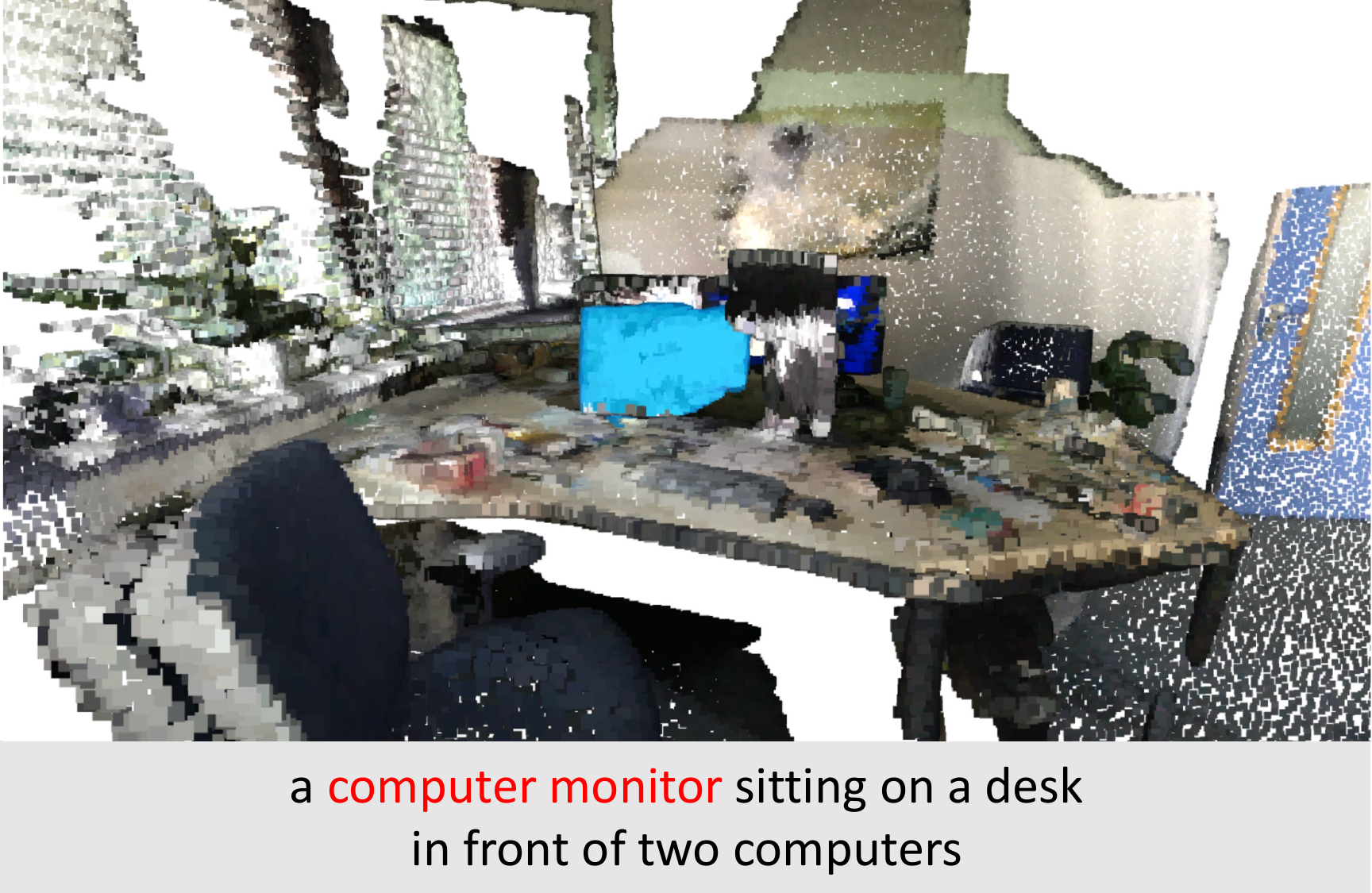}
  \end{subfigure}
  \begin{subfigure}{0.3233\linewidth}\includegraphics[width=1.0\linewidth]{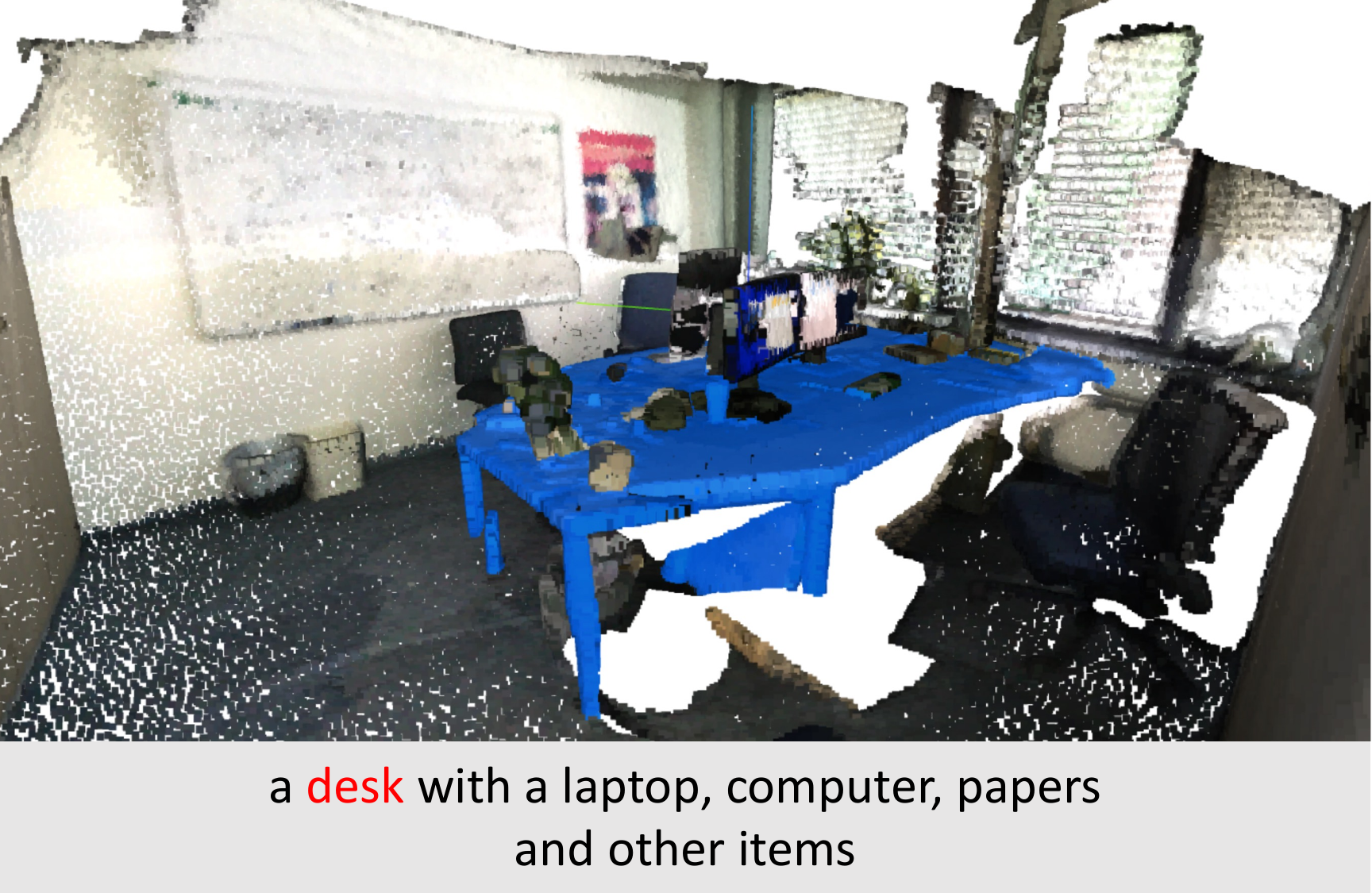}
  \end{subfigure}
  \begin{subfigure}{0.3233\linewidth}\includegraphics[width=1.0\linewidth]{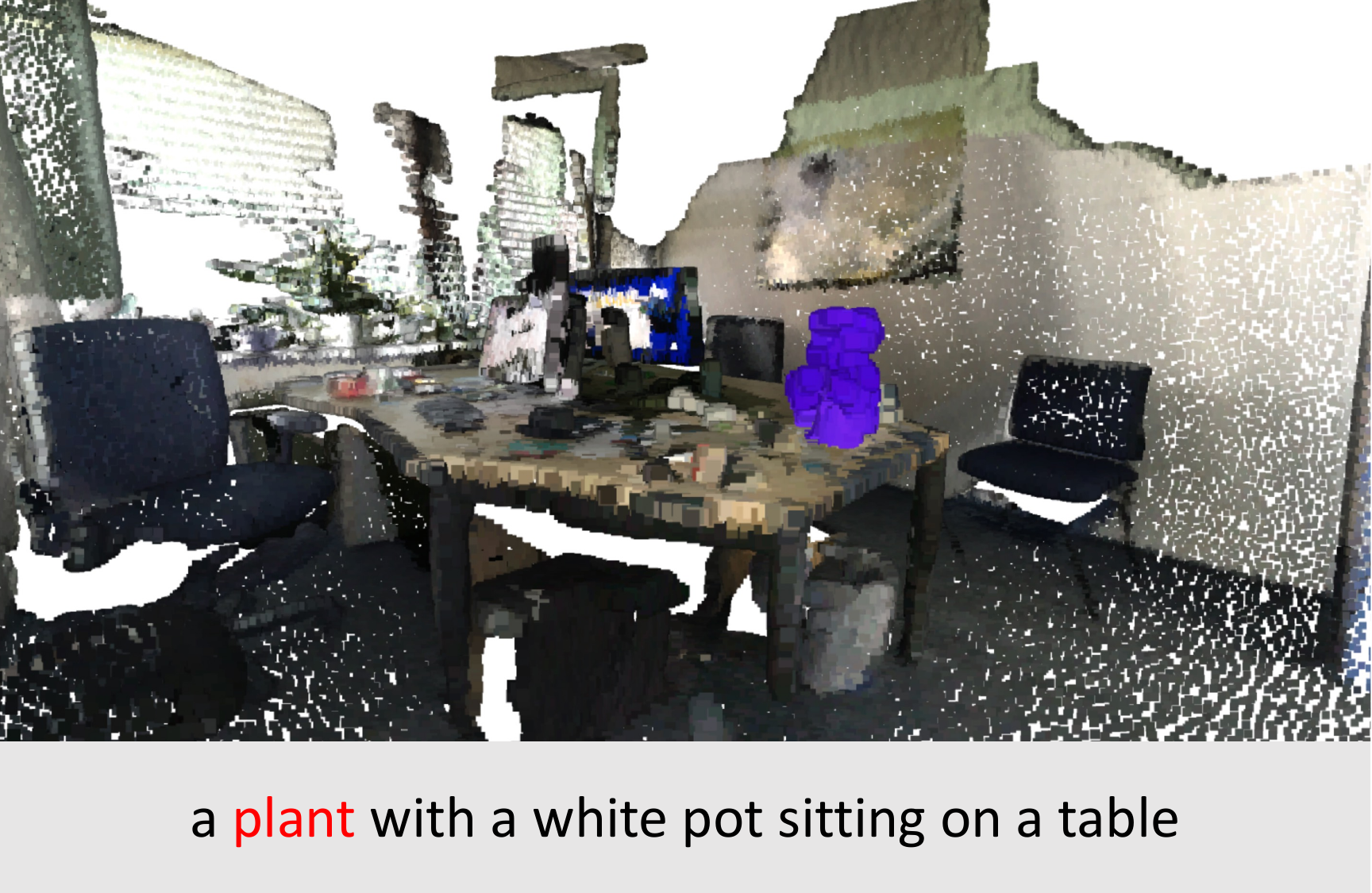}
  \end{subfigure}
  \vspace{-.1in}
  \caption{\textbf{Qualitative examples for mask captions.} Generated captions contain the semantic, appearance, and geometric relationship of the corresponding object.}
  \label{fig:caption}
  % \vspace{-20pt}
\end{figure}

As shown in Tab.~5 of the main paper, solely using 3D instance backbone feature $\mathbf{f}^{b}$ or projected CLIP visual feature $\mathbf{f}^{p}$ cannot achieve the best performance. 3D backbone features lack the generalized semantic information while projected CLIP visual features lack the location and geometry information.
To further verify the effectiveness of the backbone feature ensemble, we visualize the clustering results of different features in Fig.~\ref{fig:kmean}.
In Fig.~\ref{fig:kmean-a}, all chairs are clustered together (the green cluster), showing that the projected CLIP features contain good semantic information but cannot detect the instances. 
In Fig.~\ref{fig:kmean-b}, different instances within one category can be identified, \eg, chairs are in three clusters. However, the semantic generalization ability is degraded. As in the highlighted red circle, the trash can is detected by projected CLIP visual features (Fig.~\ref{fig:kmean-a}) but misclassified to a chair cluster when only using 3D backbone (Fig.~\ref{fig:kmean-b}).
Compared with using the two features separately, \ours combines the two features and thus achieves better semantic generalization ability (segmentation of the trash can) and segmentation performance (chairs are clustered into six clusters). The visualization results further demonstrate the effectiveness of the backbone feature ensemble.

\subsection{Inference Ensemble}
During inference, we combine the CLIP visual features with the predicted mask feature to achieve better generalization ability. Specifically, after obtaining the 3D masks, per-point CLIP features are pooled within the mask. The pooled CLIP feature and mask feature are {then} fed into the classifier to obtain the respective classification probability $p(\mathbf{f}^m)$ and $p(\mathbf{f}^p)$, and the final probability is yielded by the ensemble of them. In Tab.~\ref{tab:logit}, we compare three options to ensemble the class probabilities of 3D segmentation model and CLIP model. ``\textit{w.o. Ensemble}'' denotes only using the prediction of 3D segmentation model, and ``\textit{hard geometric mean}'' refers to the standard geometric mean, formulated as $p(\mathbf{f}^m)^{\tau} \cdot p(\mathbf{f}^p)^{1-\tau}$. Our method, ``\textit{soft geometric mean}'' (Eq.~\ref{eq:soft_geometric_mean}), shows the best results among the ensemble methods, demonstrating the effectiveness of dynamically fusing the prediction from two models. However, as shown in the first row, \ours already achieves competitive performance even without utilizing the CLIP prediction, further demonstrating the strong generalization ability of our multimodal fusion network.

\section{Qualitative Results}
\label{sec:quality}

\noindent \textbf{Visualization for Mask Captions.}
We provide the visualization for different masks and corresponding generated captions in Fig.~\ref{fig:caption}. The generated caption contains the semantic information of the 3D object as well as the location for better 3D segmentation. As shown in the examples, more than one nouns exist in the caption and thus we aggregate all the noun phrases with attention mechanism in mask-entity association.

\vspace{2mm}
\noindent \textbf{Visualization for Segmentation Results.}
More qualitative segmentation results are shown in the \href{https://cvrp-sole.github.io}{project page}. Our \ours can effectively respond to various free-form language instructions.

\bibliographystyle{splncs04}
\bibliography{egbib}
\end{document}